\documentclass[letterpaper]{article} 
\usepackage{aaai25}  
\usepackage{times}  
\usepackage{helvet}  
\usepackage{courier}  
\usepackage[hyphens]{url}  
\usepackage{graphicx} 
\usepackage{xcolor}
\usepackage{natbib}  
\usepackage{caption} 
\usepackage{algorithm}
\usepackage{algorithmic}

\usepackage{newfloat}
\usepackage{listings}
\DeclareCaptionStyle{ruled}{labelfont=normalfont,labelsep=colon,strut=off} 
\floatstyle{ruled}
\newfloat{listing}{tb}{lst}{}
\floatname{listing}{Listing}
\usepackage{amsmath}
\DeclareMathOperator{\sign}{sign}
\usepackage{xspace}
\usepackage[nameinlink,noabbrev,capitalize]{cleveref}
\newcommand{\dlgnmodel}{DLGN \xspace}
\usepackage{subcaption}
\usepackage{multicol}
\usepackage{multirow}
\usepackage{tabularx}
\usepackage{booktabs}
\usepackage{amssymb}

\crefname{appendixfigure}{Appendix Figure}{Appendix Figures}
\crefname{appendixtable}{Appendix Table}{Appendix Tables}
\crefname{appendixalg}{Appendix Algorithm}{Appendix Algorithms}
\crefname{appendixeq}{Appendix Equation}{Appendix Equations}

\crefformat{appendixfigure}{#2Appendix Figure~#1#3}
\crefformat{appendixtable}{#2Appendix Table~#1#3}
\crefformat{appendixalg}{#2Appendix Algorithm~#1#3}
\crefformat{appendixeq}{#2Appendix Equation~#1#3}

\title{Interpreting Adversarial Attacks and Defences using Architectures with Enhanced Interpretability}
\author{
    Akshay G Rao, Chandrashekhar Lakshminarayanan \& Arun Rajkumar
}

\affiliations{

    Department of Computer Science, Indian Institute of Technology Madras\\
    \texttt{akshaygrao@outlook.com, chandrashekar@cse.iitm.ac.in, arunr@cse.iitm.ac.in}
}

\begin{document}

\maketitle

\begin{abstract}
Adversarial attacks in deep learning represent a significant threat to the integrity and reliability of machine learning models. Adversarial training has been a popular defence technique against these adversarial attacks. 
In this work, we capitalize on a network architecture, namely Deep Linearly Gated Networks (\dlgnmodel), which has better interpretation capabilities than regular deep network architectures. Using this architecture, we interpret robust models trained using PGD adversarial training and compare them with standard training. Feature networks in DLGN act as feature extractors, making them the only medium through which an adversary can attack the model. We analyze the feature network of DLGN with fully connected layers with respect to properties like alignment of the hyperplanes,  hyperplane relation with PCA, and sub-network overlap among classes and compare these properties between robust and standard models. 
We also consider this architecture having CNN layers wherein we qualitatively (using visualizations) and quantitatively contrast gating patterns between robust and standard models. We uncover insights into hyperplanes resembling principal components in PGD-AT and STD-TR models, with PGD-AT hyperplanes aligned farther from the data points. We use path activity analysis to show that PGD-AT models create diverse, non-overlapping active subnetworks across classes, preventing attack-induced gating overlaps. Our visualization ideas show the nature of representations learnt by PGD-AT and STD-TR models.

\end{abstract}

\section{Introduction and related works}
\textbf{Adversarial attacks and defences: }Though machine learning algorithms perform well under normal conditions, they can fail with cleverly crafted inputs called adversarial samples, raising security concerns in many applications. White-box attacks are attacks wherein the attacker can access model predictions, parameters and training data. Popular attacks in this setting are BIM(\citet{bim_attack_paper}), MIM(\citet{mim_attack_paper}), FGSM(\citet{fgsm_paper}) and Projected Gradient Descent (henceforth abbreviated as PGD)(\citet{pgd_at_paper}) among which PGD attacks are considered one of the strongest white-box attacks in practice. Prior works have proposed various defence techniques against adversarial attacks, among which the seminal work of \citet{pgd_at_paper} stands out as one of the principled methods. They view defending adversarial attacks as solving a min-max optimization problem wherein the inner maximization aims to get the best possible adversarial samples at a given model state. They solve the inner maximization by using the PGD attack and call it \emph{adversarial training} (\cref{alg:pgd_at}) (abbreviated as PGD-AT henceforth). 

\textbf{Interpretation of robustness: } The arms race between adversarial attacks and defences has also lead to many works which instead analyse the adversarial attacks in several ways like distribution shift analysis (\citet{analysis_robustness_via_filters}), feature representation analysis by ideas like inversion (\citet{featvis_robustness_1},\citet{featvis_robustness_2}), Fourier spectrum analysis (\citet{analysis_frequencybias},\citet{analysis_LFC_effectiveness}, \citet{analysis_robustness_fourier_perspective}, \citet{analysis_robustness_hfc_help_generalization}, \citet{analysis_spectralbiasneuralnetworks}), principal component analysis (\citet{analysis_robustness_via_filters}), shapely value analysis (\citet{analysis_robustness_frequency_shapley}).

\begin{figure}
    \centering
    \includegraphics[width=\linewidth]{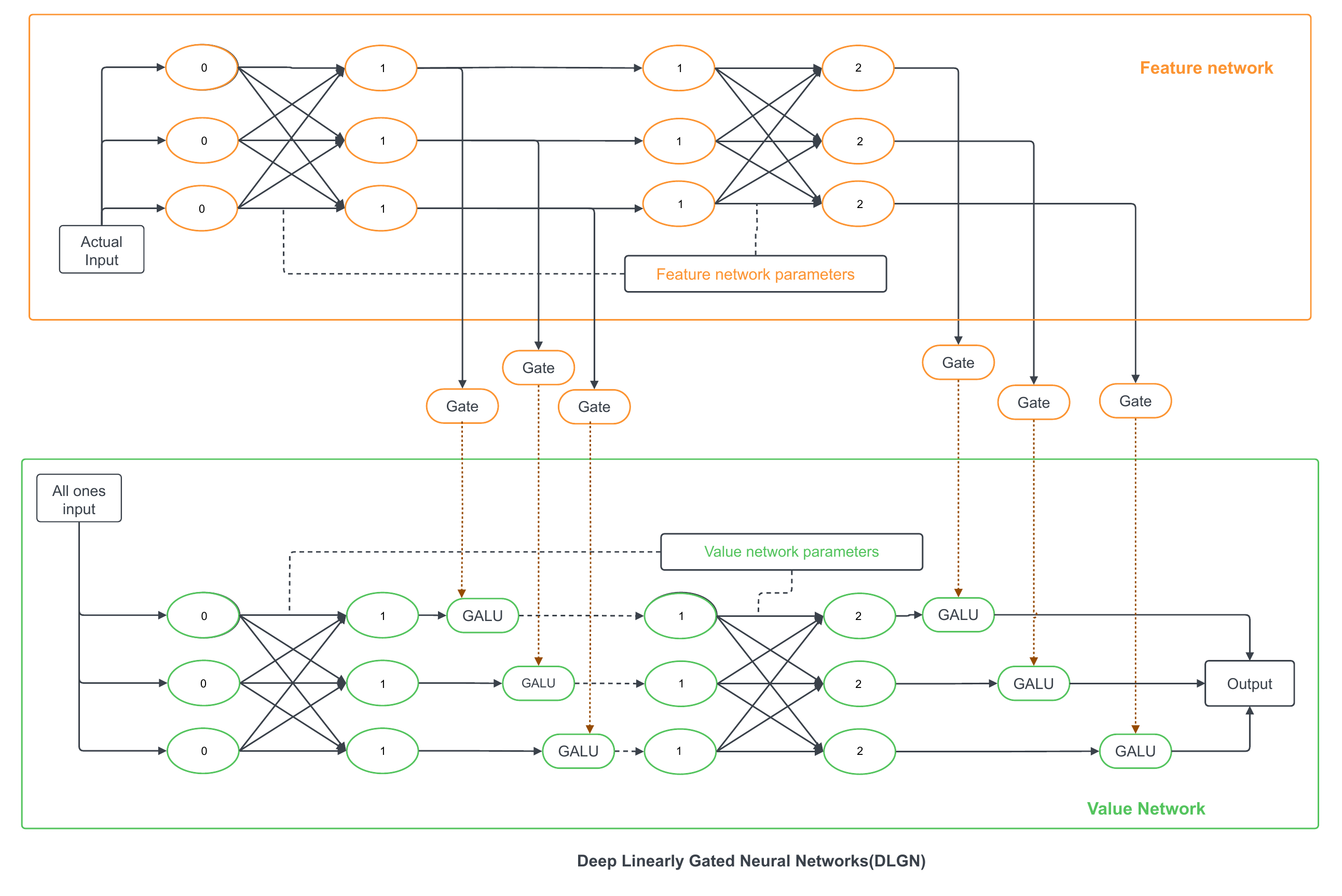}
    \caption{Deep Linearly Gated Networks (DLGN) network architecture. \(GALU=x*\mathit{Gate(x^`)}\)}
    \label{fig:appnd_dlgn_net_arch}
\end{figure}

In this work, we consider a recently proposed novel neural network architecture that is much more amenable to adversarial training analysis. 

\textbf{Deep Linearly Gated Networks: }Relu activation can be viewed as the product of input and gates that are off/on. These gates trigger certain pathways in the network to be active/inactive. \citet{neurips_role_of_gates} propose a unique approach by viewing model training as active sub-network learning in Relu-activated neural networks. Neural networks can be viewed as model input being mapped into the path space (path space representation given by neural path features (NPF)) and combined (coefficients of these combinations are given by neural path values (NPV)) in the path space to generate model output logits (Refer \Cref{eq:npf_dot_npv_proof} for more details). They introduce Deep Gated Neural Network (DGN) architecture to demonstrate the role of active sub-network learning that has two nearly identical sub-networks: \emph{feature network}, which is responsible for extracting features and providing gating signals (thereby solely encoding NPFs); \emph{value network}, which aggregates the features extracted by the feature network (thereby solely encoding NPVs) to produce the final model prediction. 

\textbf{Model with enhanced interpretability: }A follow-up study by \citet{dlgn_paper} show that interpreting the value network \emph{visually} is meaningless in DGN networks. However, interpreting feature network is still hard due to the non-linearity in the feature network layers. So, to improve the interpretability of DGNs, they propose a new architecture, namely \emph{Deep Linearly Gated Neural Networks (DLGN)} (see \Cref{fig:appnd_dlgn_net_arch} for network architecture) wherein the gating signals are completely moved out of the feature network, rendering the transformations in the feature network entirely linear. The DLGN architecture offers significant interpretability advantages compared to standard architectures due to the linear feature network.

\textbf{Goal:} 
We use the enhanced interpretation capabilities of the \dlgnmodel model to compare and contrast standard training (henceforth abbreviated as STD-TR) and adversarial training by analysing the model's internals. We use ideas like hyperplane analysis, which are unique to DLGN architecture, unlike standard architectures, and ideas from feature visualization and path-view in neural networks to give multi-dimensional insights into robustness.

\textbf{Our Contributions}
\begin{itemize}
    \item We merge layers in the feature network of \dlgnmodel architectures to obtain a single effective linear transformation per layer, which is unique to our work. This reveals novel insights into hyperplanes and their resemblance to principal components in PGD-AT and STD-TR models. Our analyses show that hyperplanes in PGD-AT (FC) models are farther from data points compared to STD-TR (FC) models and play a key role in enhancing robustness.
    \item We analyze path activity among classes by examining the active-subnetwork overlap (employing neural path kernel) in PGD-AT and STD-TR FC models. Our findings indicate that PGD-AT models generate more diverse active subnetworks and can avoid active subnetwork overlaps with different classes during an attack.
    \item We quantitatively compare active gate overlaps among classes using the intersection-over-union idea. This reveals that adversarially trained models can prevent significant gating pattern changes and avoid overlap of attack-induced gating changes with those of other classes. 
    \item Using visualization ideas, we interpret the representations used by PGD-AT and STD-TR models in the feature network of \dlgnmodel.
\end{itemize}

\section{Analysis of hyperplanes in feature network of fully connected robust and standard models}
\textbf{Notations}
Let $\theta_{f}$ and $\theta_{v}$ be parameters of the model with $L$ layers in feature network and value network respectively and more specifically with $W_{l} \in R^{m_{l-1}, m_{l}}$ being the weight at layer $l$ of feature network,  $b_{l} \in R^{m_{l}}$ being the bias at layer $l$ of the feature network. Let $x_{l} \in R^{m_{l}}$ be the feature network output at layer $l$, $p$ be one of the paths among total $P$ paths passing from each input node to each output node, $G_{x,\theta_{f}}^{l,p}$ be the gate for input $x$ at the node contained in path $p$ at layer $l$ and $x_{p}$ be the input node at node contained in path $p$. Then, from work [\citet{neurips_role_of_gates}], the gate information is encoded in the neural path features (NPF) $\Phi_{x,\theta_{f}} \in R^{P}$ (\Cref{eq:npf_definition}) as the product of input and gates along a path. Similarly, the weight information is encoded in the neural path value (NPV) $\vartheta_{\theta_{v}} \in R^{P}$ (\Cref{eq:npv_definition}) as the product of weights along a path. The final model output logit per output node is given by $\hat{y}(x)$, which can be expressed as per \Cref{eq:npf_dot_npv}. Refer \Cref{fig:path_view_neural_networks} for more details.
\begin{subequations}
    \begin{align}
    G^{l}_{x,\theta_{f}} &= \sigma(\beta* (W_{l}^{T}x_{l-1} + b_{l})) \quad\in R\\
    \Phi_{x,\theta_{f}} & = \{x_{p}\Pi_{l=1}^{L}G_{x,\theta_{f}}^{l,p}, \quad p \in [P]\} 
     \quad \in R^{P} \label{eq:npf_definition} \\
    \vartheta_{\theta_{v}} & = \{\Pi_{l=1}^{L}\theta_{v}^{l,p}, \quad p \in [P]\} 
         \quad \in R^{P} \label{eq:npv_definition}\\
    \hat{y}(x) & = <\Phi_{x,\theta_{f}},\vartheta_{\theta_{v}}> \quad \in R \label{eq:npf_dot_npv} \\
    \text{where $\sigma$} & \text{ is the sigmoid activation i.e $\sigma(x)=\frac{1}{1+e^{-x}}$} \notag
    \end{align}
\end{subequations}

Consider a \dlgnmodel architecture with fully connected layers where the feature network is entirely linear. At each  \emph{feature network layer} \( l \), the effective linear transformation can be obtained by merging all preceding layers up to \( l \), with effective weights \( E_{l} \in R^{m_{0},m_{l}} \) and bias \( p_{l} \in R^{m_{l}} \). \textbf{Note that this analysis is not possible in standard neural networks due to the non-linearity between layers}. The output at layer \( l \) would produce \( m_{l} \) gates and each gate's effective weight \(\in R^{m_{0}}\) would be a hyperplane acting on input in \( m_{0} \)-dimensional space. A gate is active/inactive based on which side of the hyperplane the input \( x \) lies.

\begin{subequations}
    \begin{align}
    \hat{y}(x+\delta) &= \sum^{P}_{p=1}{\Phi_{x+\delta,\theta_{f}} * \vartheta_{\theta_{v}}} \notag \\
     = \sum & ^{P}_{p=1}{[(x^{p}+\delta^{p}) \Pi^{L}_{l=1} \sigma\{E_{l}^{p}(x+\delta)+p_{l})\}]}*\vartheta_{\theta_{v}} \label{eq:npf_dot_npv_perturb}
    \end{align}
\end{subequations}

From \Cref{eq:npf_dot_npv_perturb} for a perturbation $\delta$ in input $x$, larger values of \( E_{l}^{p}x + p_{l} \) reduce the gate's sensitivity in path \( p \), thereby enhancing adversarial robustness. Informally, if a point is farther from a hyperplane, it requires either larger dimension-wise perturbations or small perturbations across many dimensions to flip the gate\footnote{In experiments, \( \beta \) is set high to approximate a step function}.

\subsection{Hyperplane analysis in real-world dataset}

\begin{figure}[]  
    \centering
    \begin{subfigure}[b]{0.49\linewidth}  
        \centering
        \includegraphics[width=\textwidth]{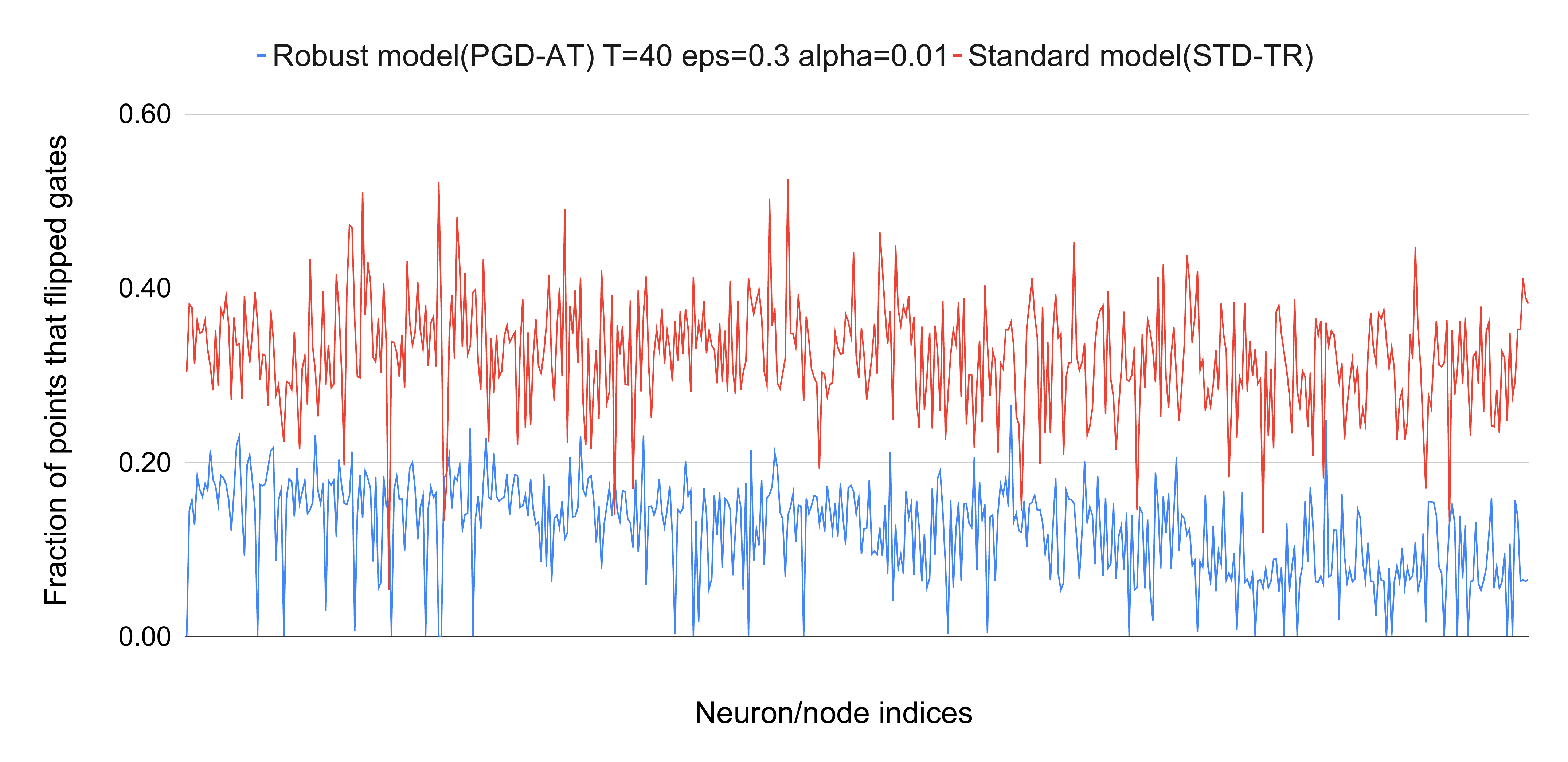}  %
        \caption{MNIST dataset.}
        \label{fig:fc_dlgn_flip_distribution_mnist}
    \end{subfigure}
    \hfill
    \begin{subfigure}[b]{0.49\linewidth}
        \centering
        \includegraphics[width=\textwidth]{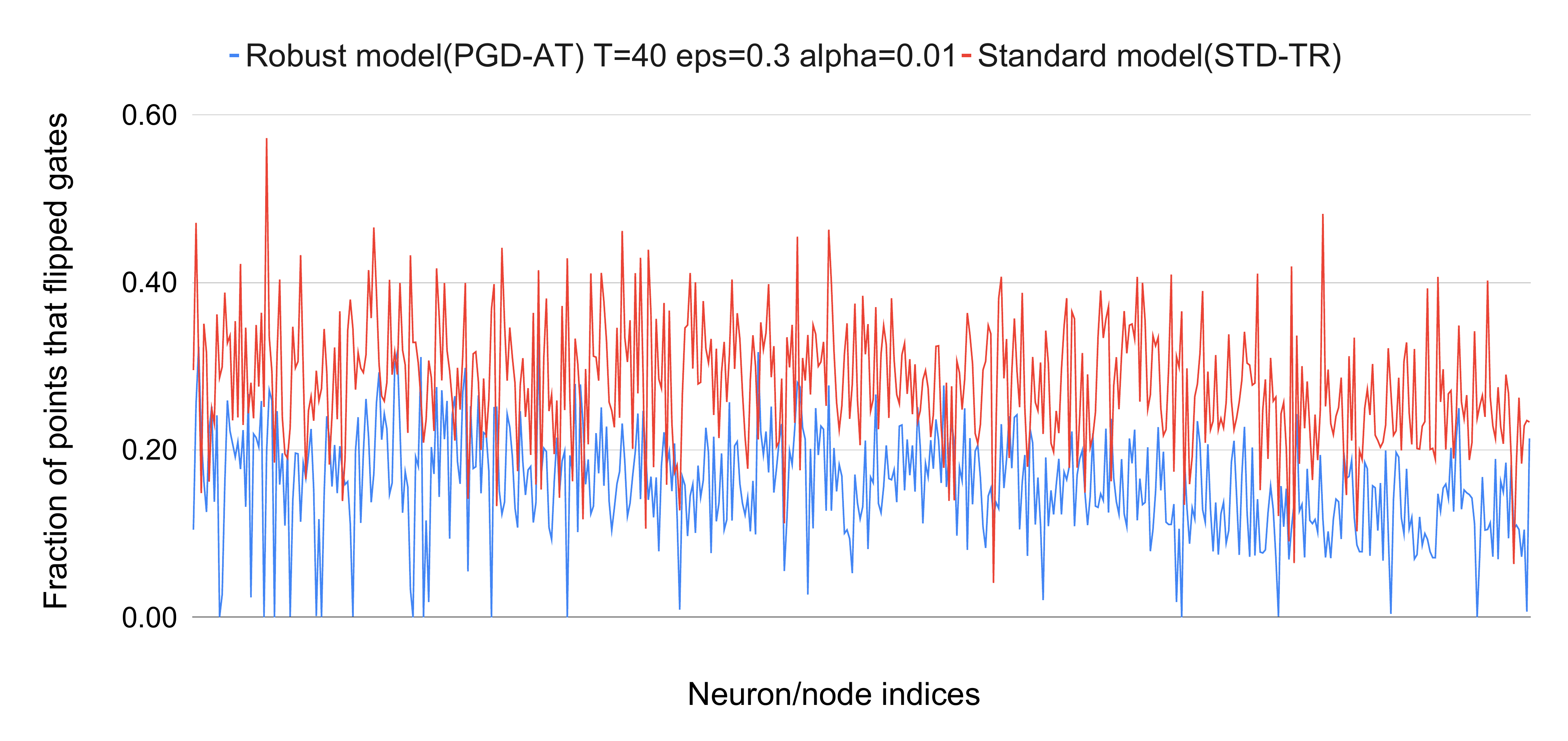}  
        \caption{Fashion MNIST dataset.}
        \label{fig:fc_dlgn_flip_distribution_famnist}
    \end{subfigure}
    \caption{PGD-AT vs STD-TR FC-\dlgnmodel-W128-D4 flip distribution. The Y-axis denotes the fraction of points that flipped the gate at node indices on the X-axis.}
    \label{fig:fc_dlgn_flip_distribution}
\end{figure}

\begin{figure}[]
    \centering
    \begin{subfigure}[c]{0.48\linewidth}
        \centering
        \includegraphics[width=\textwidth]{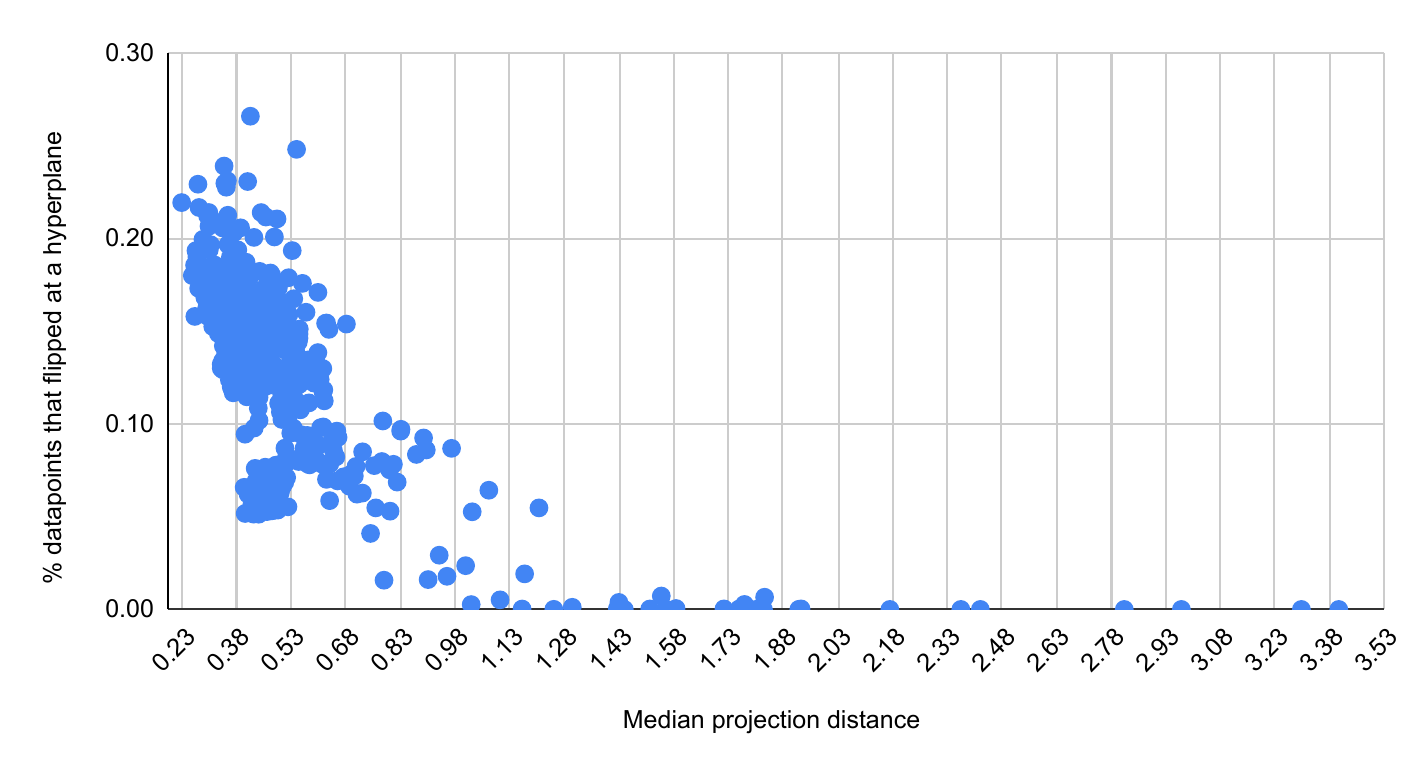}
        \caption{PGD-AT model.}
        \label{fig:fc_dlgn_pgdat_flip_dist_vs_median_dist}
    \end{subfigure}
    \hfill
    \begin{subfigure}[c]{0.48\linewidth}
        \centering
        \includegraphics[width=\textwidth]{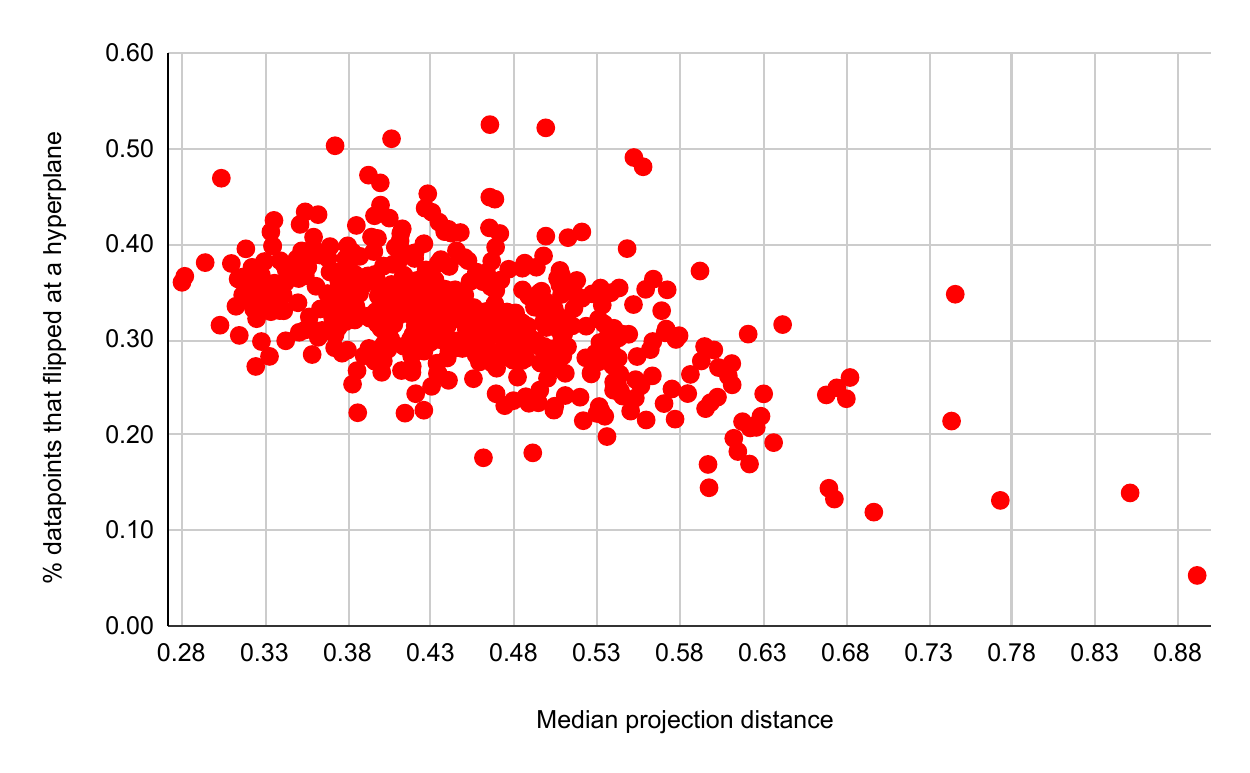}
        \caption{STD-TR model.}
        \label{fig:fc_dlgn_stdtr_flip_dist_vs_median_dist}
    \end{subfigure}
    \caption{Flip distribution per hyperplane(y-axis) vs. median projection distance(x-axis) in MNIST dataset. Each point indicates a hyperplane.}
    \label{fig:fc_dlgn_flip_dist_vs_median_dist}
\end{figure}

\begin{figure}
\begin{subfigure}[c]{0.48\linewidth}
  \includegraphics[width=\linewidth]{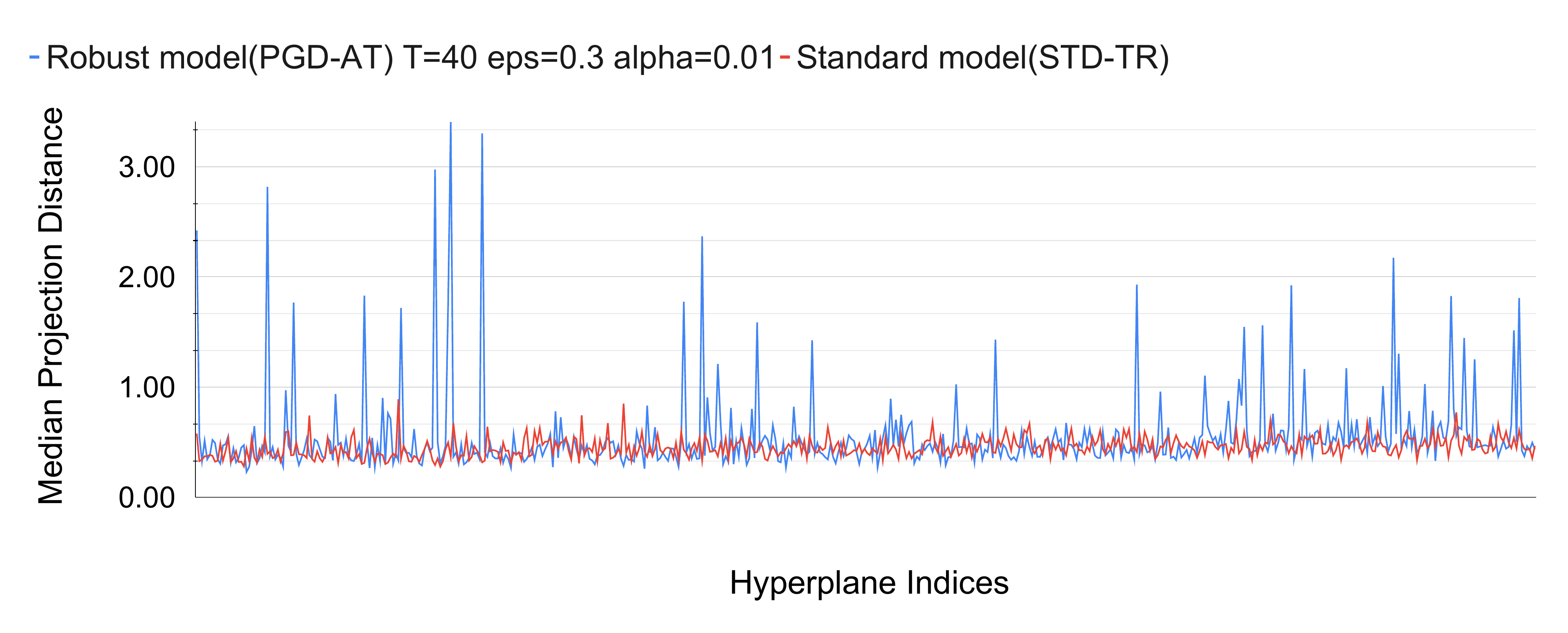}
\end{subfigure}
\begin{subfigure}[c]{0.48\linewidth}
\includegraphics[width=\linewidth]{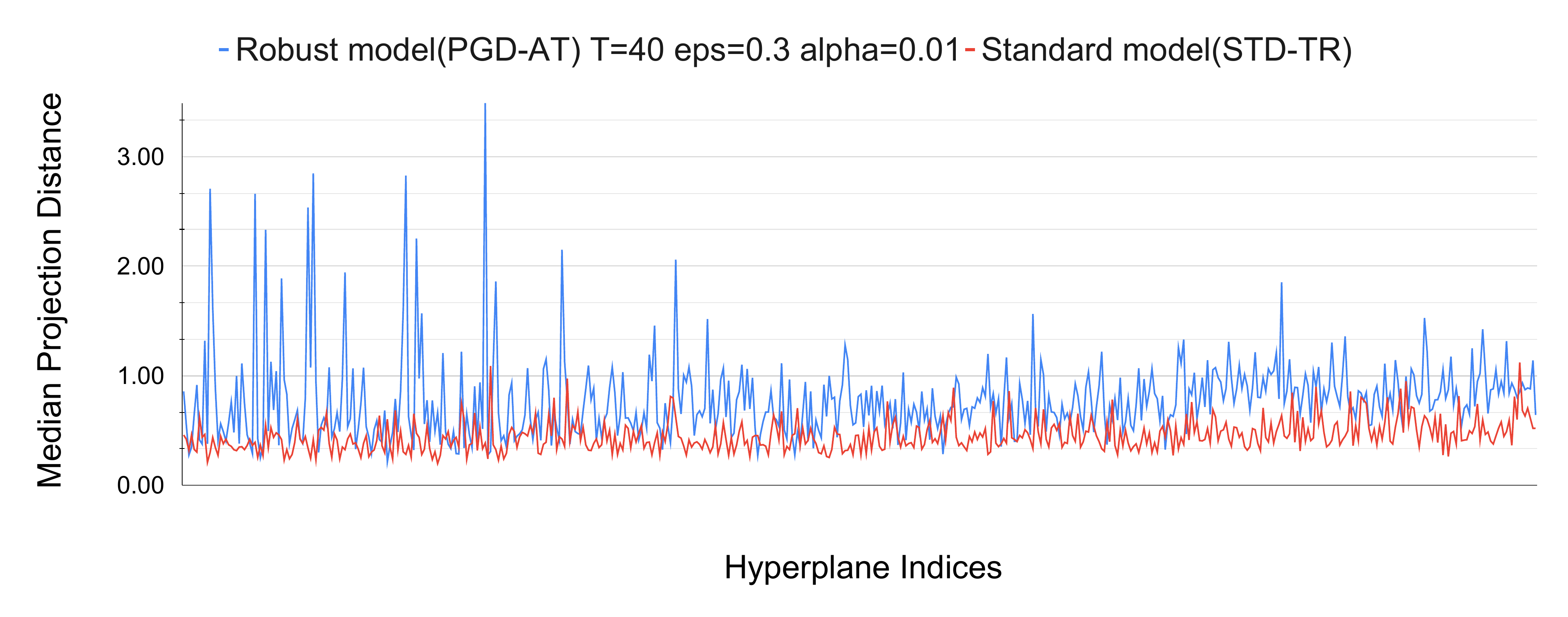}
\end{subfigure}
\caption{PGD-AT vs STD-TR FC-\dlgnmodel-W128-D4 median projection distance. The left image is on MNIST, and the right image is on the Fashion MNIST dataset. The Y-axis denotes the median projection distance of data points at node/hyperplane indices on the X-axis.}
\label{fig:fc_dlgn_median_distance_from_HP}
\end{figure}

\begin{figure}
    \centering
    \includegraphics[width=\linewidth,keepaspectratio]{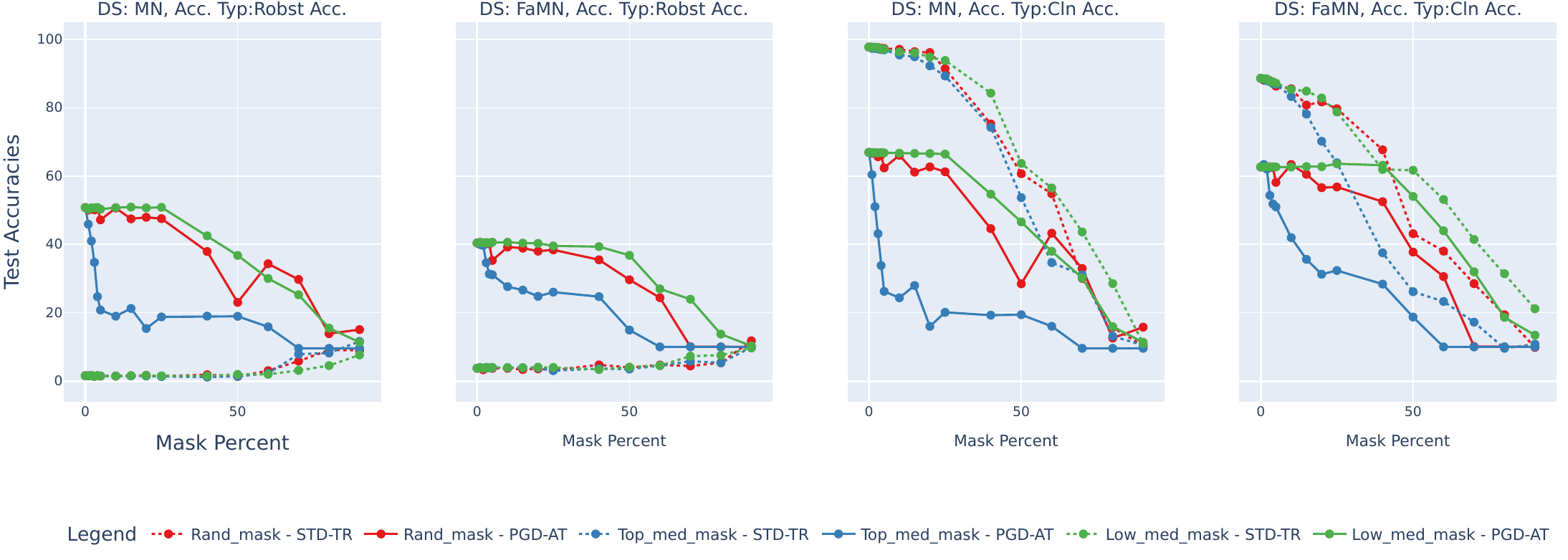}
    \caption{Robust and clean accuracies of PGD-AT and STD-TR FC-\dlgnmodel\_W128\_D4 models with random gate masking vs. masking gates with the highest median projection distance vs masking gates with lowest median projection distance. Dotted lines are for STD-TR models and solid lines are for PGD-AT models.}
    \label{fig:fc_dlgn_masking}
\end{figure}

We trained a \dlgnmodel with four fully connected layers (width 128) on the MNIST and Fashion MNIST datasets using both standard training and adversarial training (PGD-AT, \( \epsilon = 0.3, \alpha = 0.1, T = 40 \)). Adversarial attacks used PGD with 40 steps and \( \epsilon = 0.3 \). When an adversarial example crosses to the opposite side of the hyperplane compared to the original input, the gate is considered flipped (from active to inactive or vice versa). As shown in \Cref{fig:fc_dlgn_flip_distribution}, fewer data points flipped at each hyperplane in PGD-AT models than in STD-TR models. We inspect the projection distance of points from each hyperplane of fully connected layers in the feature network of \dlgnmodel given by the expression $\frac{E_{l}^{T}x+p_{l}}{||E_{l}||_{2}}$. Guided by the mathematical intuition at \Cref{eq:npf_dot_npv_perturb}, we experimentally (in \Cref{fig:fc_dlgn_pgdat_flip_dist_vs_median_dist,fig:fc_dlgn_stdtr_flip_dist_vs_median_dist}) show that larger median projection distances results in less gate flipping thereby enhancing robustness. We plot the median projection distance over all samples from each hyperplane across all layers (see \Cref{fig:fc_dlgn_median_distance_from_HP} and \cref{fig:appnd_fc_dlgn_median_distance_from_HP_256}) and found that the median distance from hyperplane is relatively higher in PGD-AT models than STD-TR models at many hyperplane indices. This trend is also reflected in projection distance histograms, which show significant differences between standard and robust models (see \Cref{fig:appnd_projection_distance_distribution}). In \Cref{fig:fc_dlgn_masking}, we compare masking gates with the highest median projection distance, masking gates with the lowest median projection distance and masking gates randomly in PGD-AT and STD-TR models. Results show that median-based masking significantly reduces PGD-40 and clean accuracies in PGD-AT models, highlighting the importance of gates with higher median distances for robustness.

\subsection{Hyperplane analysis in synthetic XOR dataset}

\begin{figure*}[t]
  \centering
    \begin{subfigure}[b]{0.25\textwidth}
    \includegraphics[width=\textwidth,keepaspectratio]{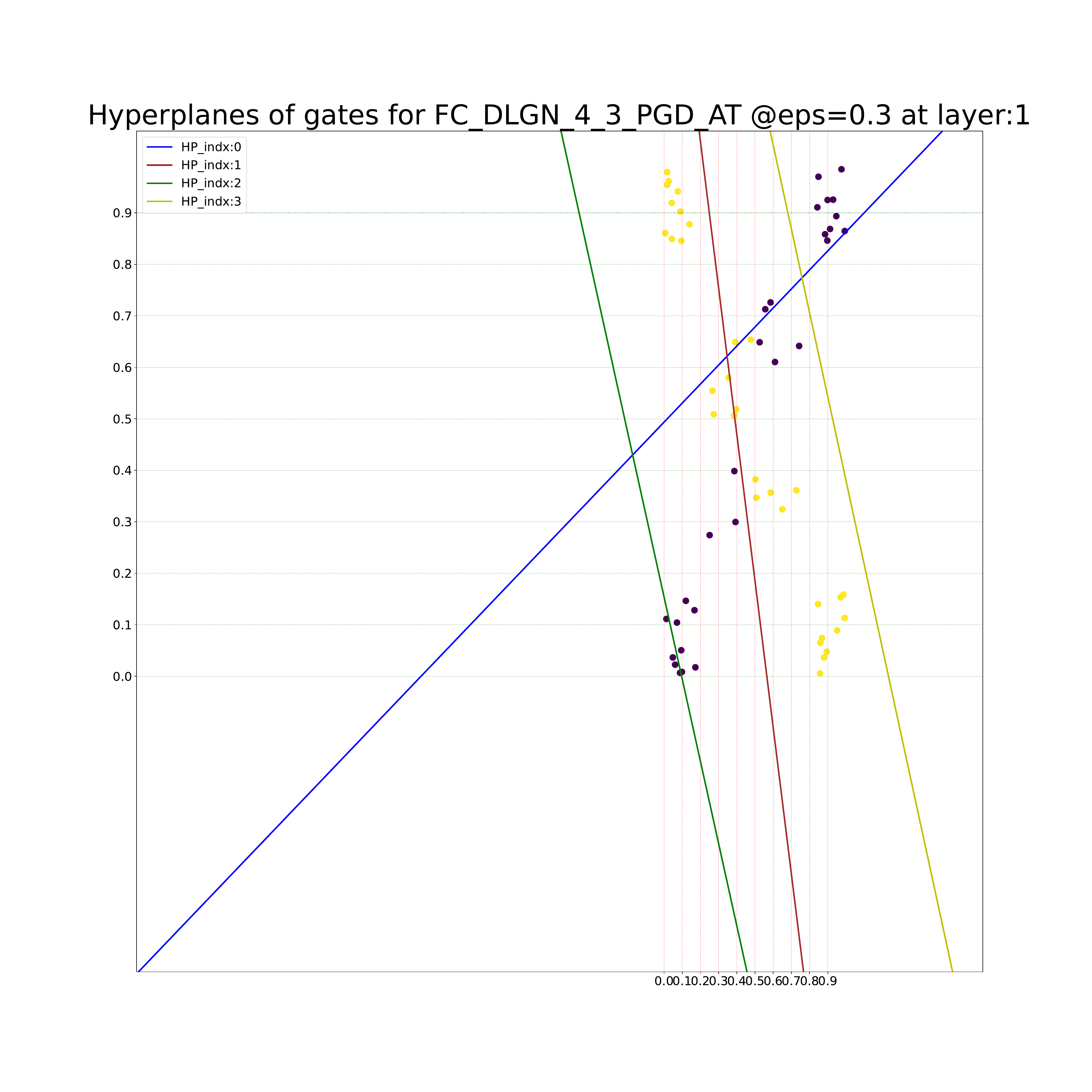}
    \label{fig2:xor_hyperplanes_pgd1}
    \end{subfigure}
    \begin{subfigure}[b]{0.25\textwidth}
    \includegraphics[width=\textwidth,keepaspectratio]{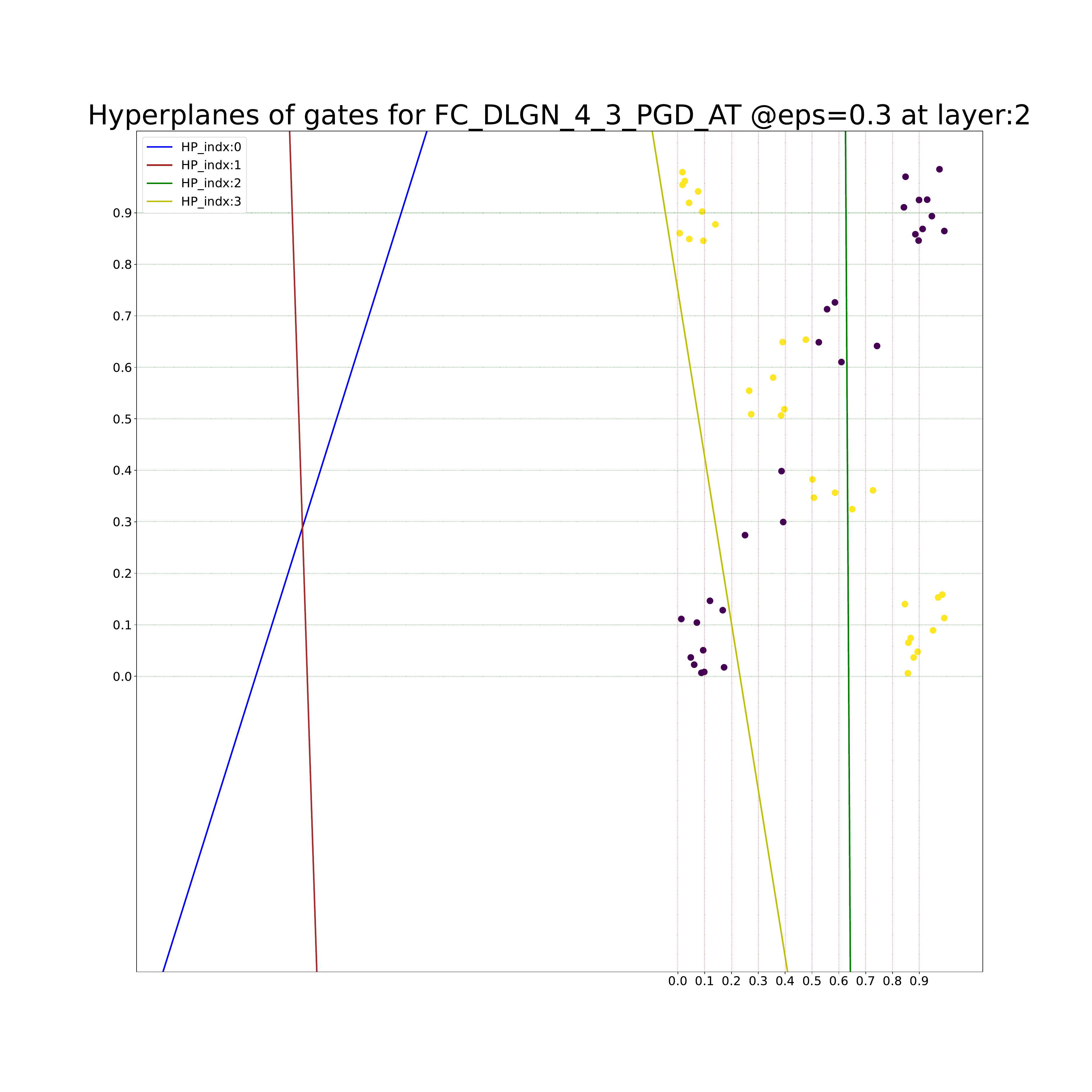}
    \label{fig2:xor_hyperplanes_pgd2}
    \end{subfigure}
    \begin{subfigure}[b]{0.25\textwidth}
    \includegraphics[width=\textwidth,keepaspectratio]{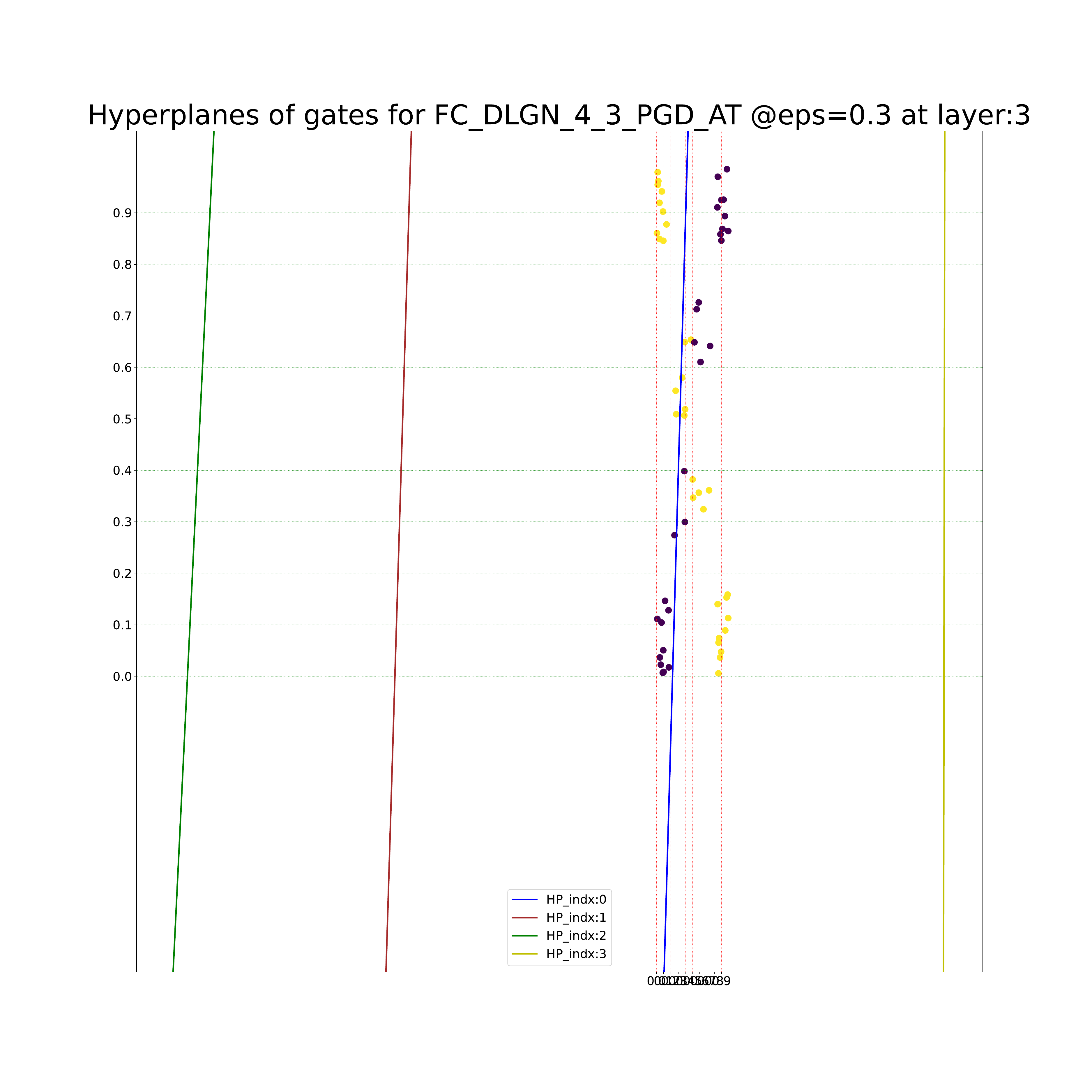}
    \label{fig2:xor_hyperplanes_pgd3}
    \end{subfigure}
    \begin{subfigure}[b]{0.25\textwidth}
    \includegraphics[width=\textwidth,keepaspectratio]{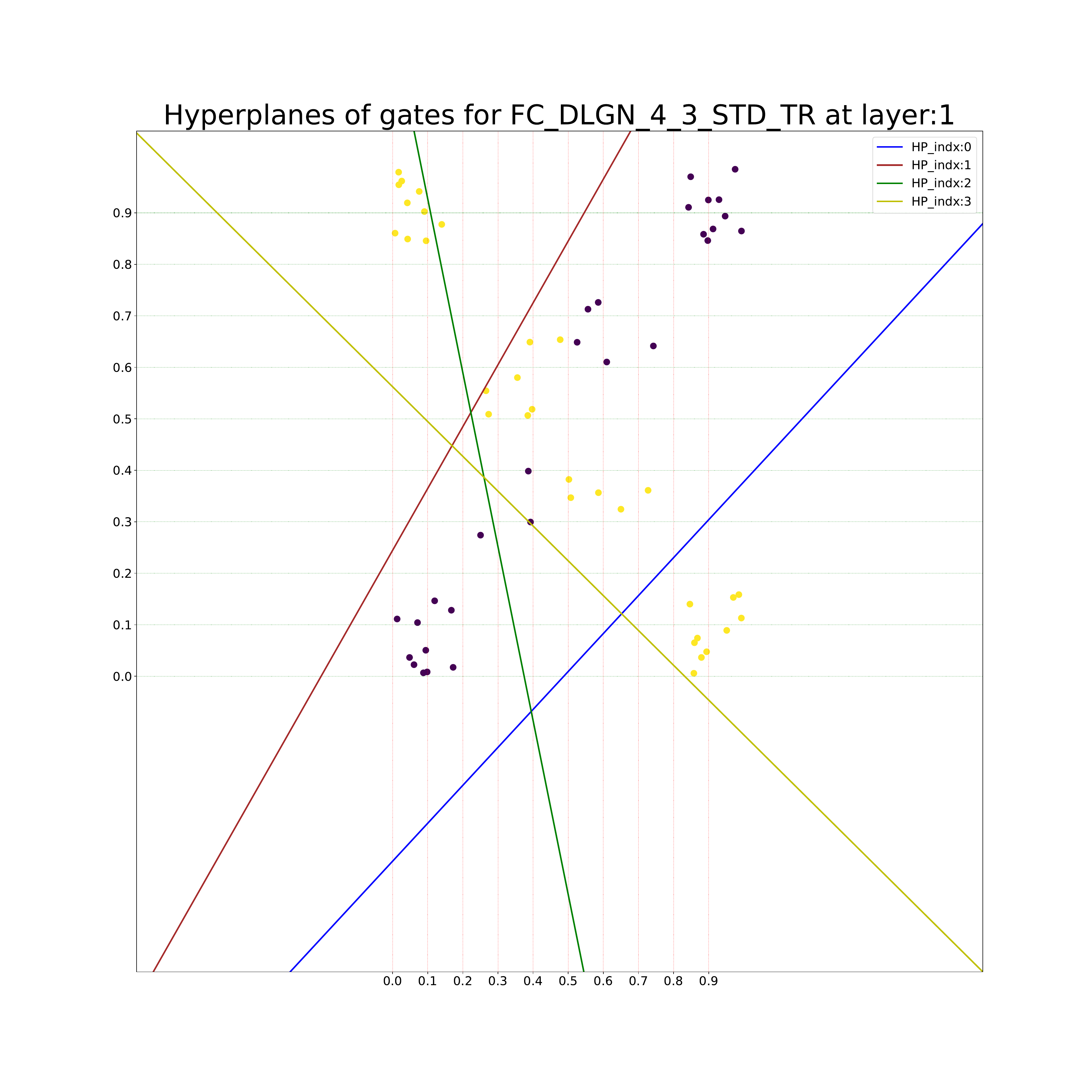}
    \label{fig2:xor_hyperplanes_std1}
    \end{subfigure}
    \begin{subfigure}[b]{0.25\textwidth}
    \includegraphics[width=\textwidth,keepaspectratio]{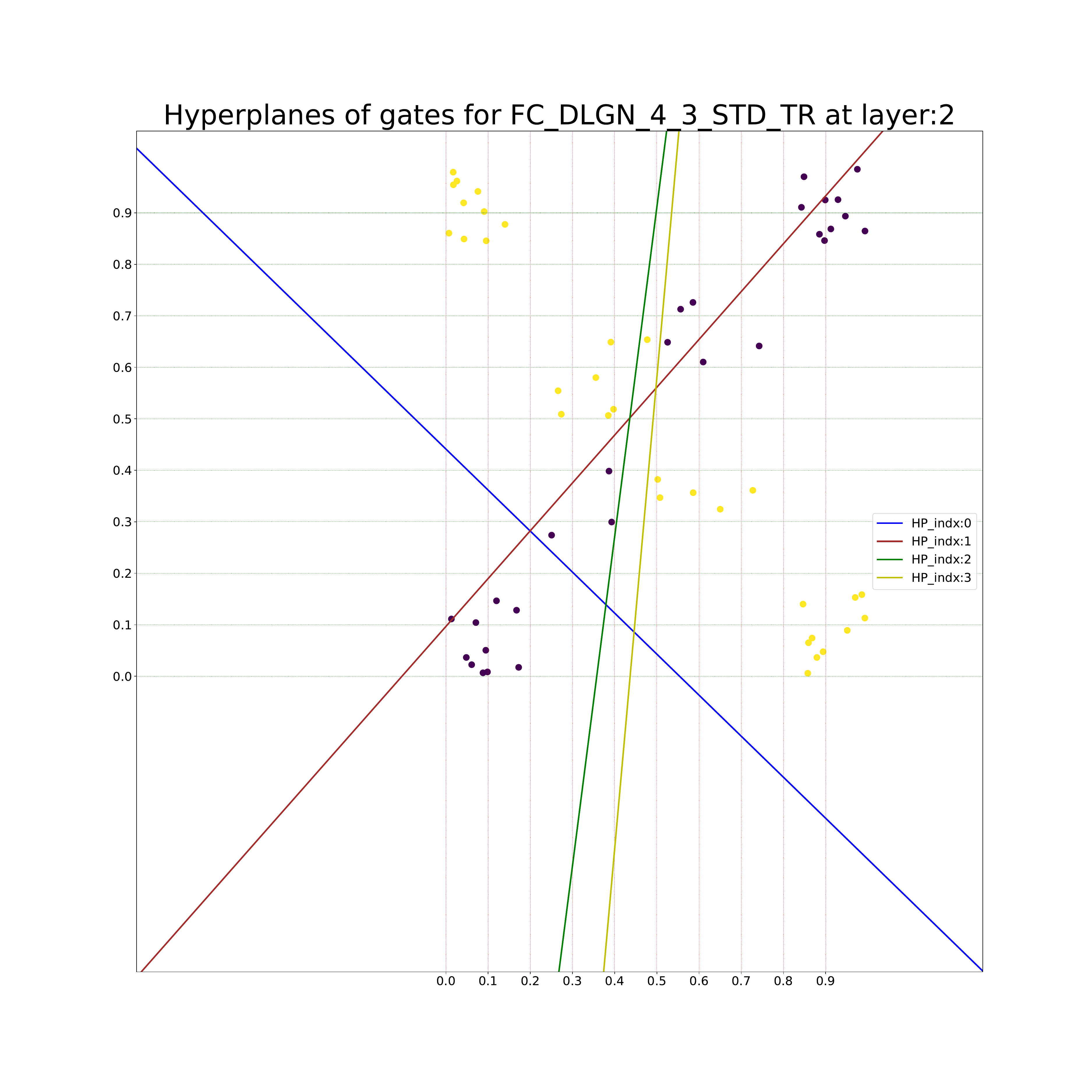}
    \label{fig2:xor_hyperplanes_std2}
    \end{subfigure}
    \begin{subfigure}[b]{0.25\textwidth}
    \includegraphics[width=\textwidth]{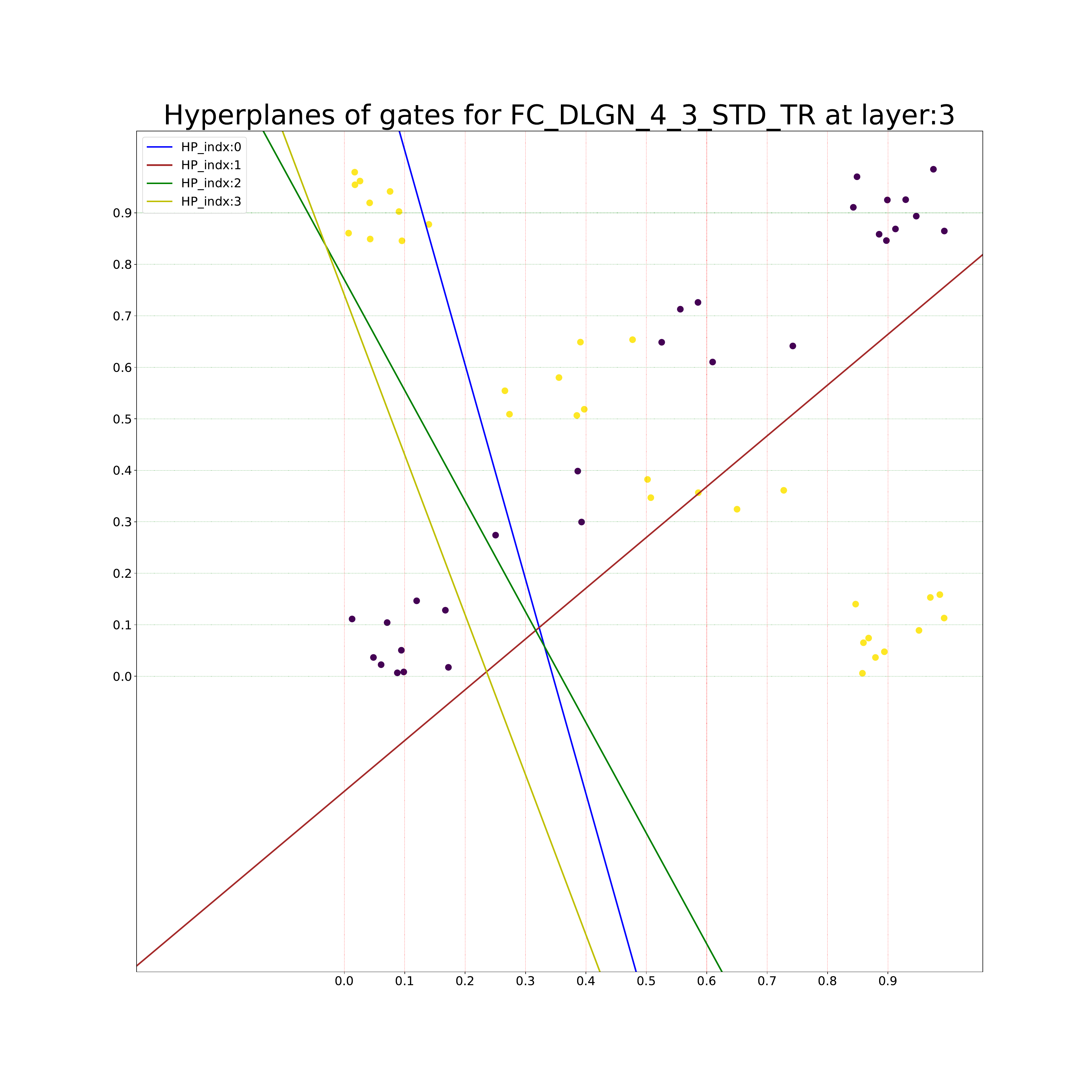}
    \label{fig2:xor_hyperplanes_std3}
    \end{subfigure}
    \caption{Hyperplane plots of PGD-AT vs STD-TR models in FC-\dlgnmodel-W4-D3. Row 1 indicates the PGD-AT model, and row 2 indicates the STD-TR model. Columns 1-3 indicate layers 1-3. Each image contains four hyperplanes since the width at each layer is 4.}
    \label{fig:xor_hyperplanes}
\end{figure*}

We constructed a 2D XOR dataset (see \Cref{fig:xor_dataset_train}) with a gap \( \lambda \) from the axes, ensuring that points satisfy \( |x| > \lambda \) and \( |y| > \lambda \). This design allows setting \( \epsilon \leq \lambda \) during adversarial training, where \( \epsilon \) represents the perturbation boundary without changing the ground truth labels. Using a \dlgnmodel with 3 fully connected layers (width 4), we trained models via both standard (STD-TR) and adversarial training (PGD-AT, \( \epsilon = 0.3, T = 40 \)). The decision boundaries of PGD-AT models are closer to optimal compared to STD-TR (see \Cref{fig:xor_decision_boundary_pgdat,fig:xor_decision_boundary_stdtr}), ensuring that adversarial examples within \( L_\infty \) bounds (\( \epsilon = 0.3 \)) are correctly classified only by PGD-AT. Visualization of hyperplanes at each layer of the feature network (see \Cref{fig:xor_hyperplanes}) shows that PGD-AT models learn hyperplanes positioned farther from the data points than STD-TR models. This trend increases in deeper layers as compared to earlier ones. So, we conclude that hyperplanes with larger projection distances from data points are key in enhancing robustness.

\section{PCA analysis in robust and standard models}
\begin{figure}
    \centering
    \includegraphics[width=0.4\textwidth,keepaspectratio]{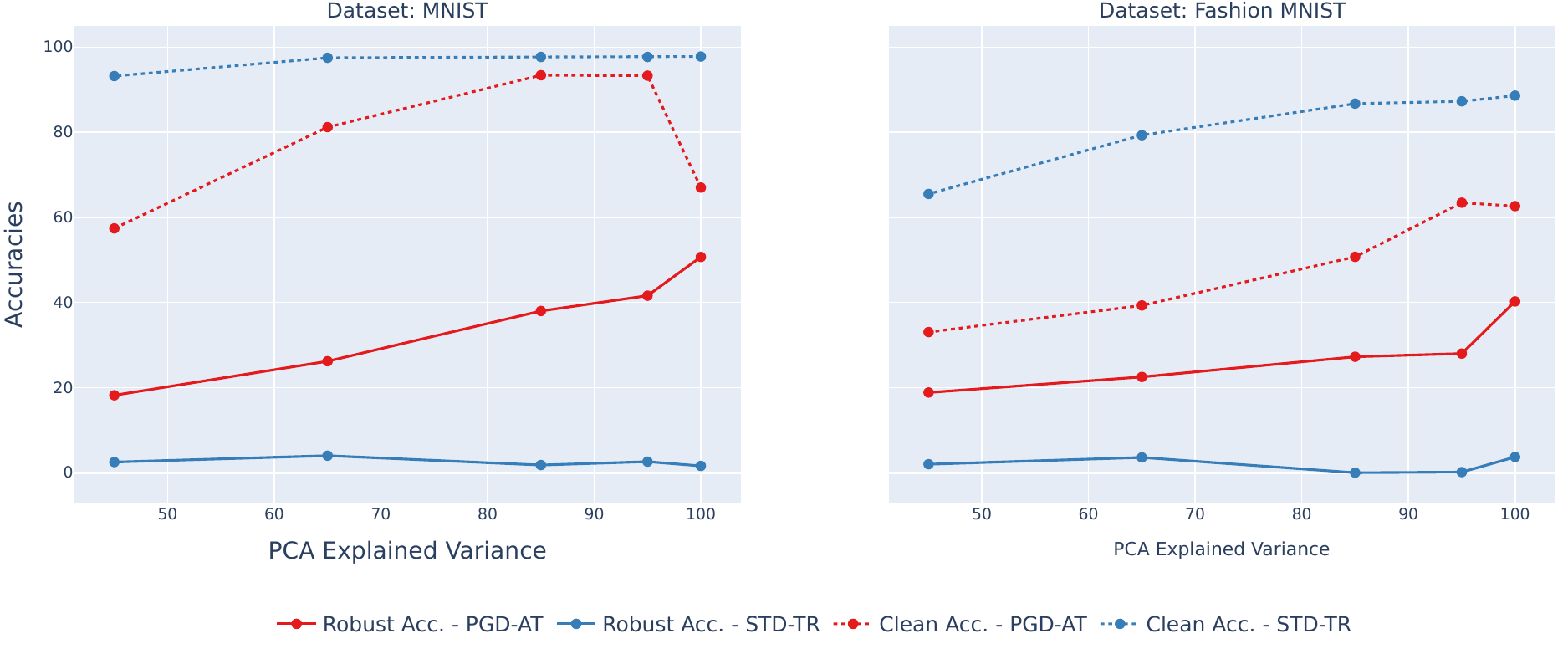}
    \caption{\dlgnmodel model trained with PCA embedded layer at different levels of dimensionality reduction on MNIST and Fashion MNIST datasets.}
    \label{fig:fc_dlgn_pca_same_capacity}
\end{figure}

Principal Component Analysis (PCA) minimizes point-to-hyperplane distances, while we saw that the PGD-AT process increases these distances to improve robustness. This fundamental difference motivates us to investigate the impact of PCA on adversarial training.
We \emph{embedded} PCA projection operation into the input layer of a \dlgnmodel architecture, ensuring both training and inference accounted for the transformation. This also ensures that the adversary has knowledge of the operation and doesn't change the dimensions of the model input. To offset the reduced capacity from PCA's dimensionality reduction, we increased the model's width at all layers to keep the capacity constant across all models under comparision. Experiments on MNIST and Fashion MNIST (see \Cref{fig:fc_dlgn_pca_same_capacity}) reveal a significant drop in both PGD-40 and clean accuracy in PGD-AT models compared to STD-TR models, indicating that PCA negatively affects adversarial robustness. This suggests that PCA's dimensionality reduction conflicts with the robustness objectives of adversarial training.

\begin{figure}
  \centering
    \begin{subfigure}[b]{0.99\linewidth}
    \includegraphics[width=\textwidth]{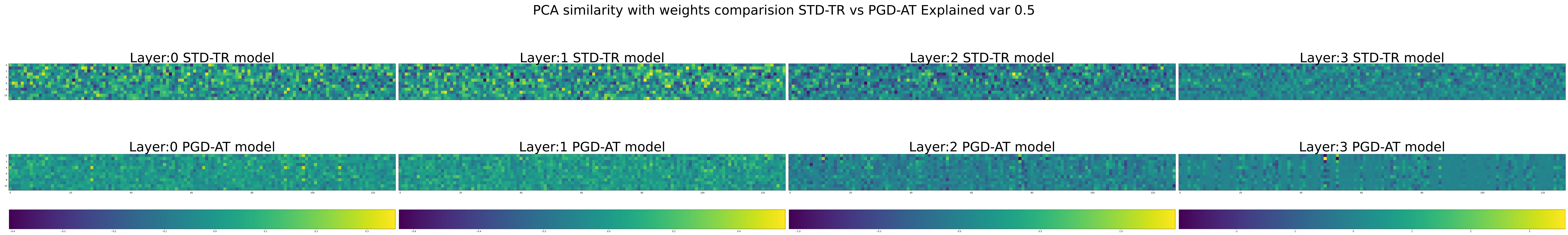}
       \caption{MNIST dataset with explained variance $0.5$, $12$ components.}
        \label{fig:fc_pca_components_corr_wrt_weights_mnist}
    \end{subfigure}
    \begin{subfigure}[b]{0.99\linewidth}
    \includegraphics[width=\textwidth]{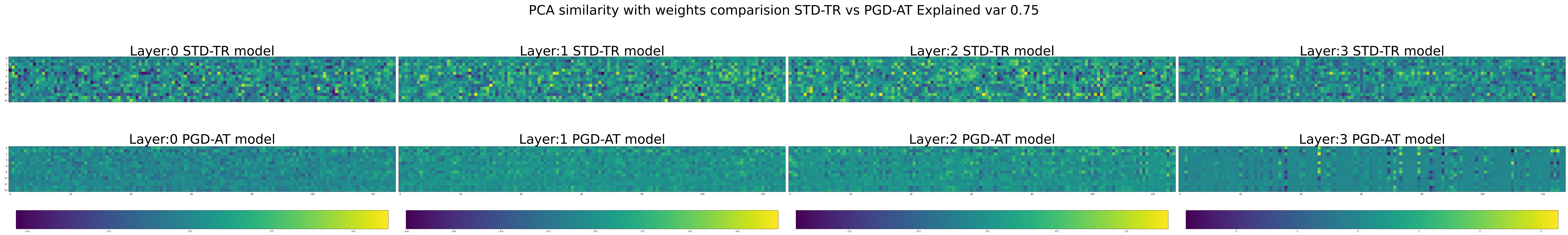}
       \caption{Fashion MNIST dataset with explained variance $0.75$, $15$ components.}
        \label{fig:fc_pca_components_corr_wrt_weights_famnist}
    \end{subfigure}
    \caption{Effective weights with top PCA components in PGD-AT(bottom row) and STD-TR(top row) using FC-\dlgnmodel-W128-D4 architecture.}
    \label{fig:fc_pca_components_corr_wrt_weights}
\end{figure}

To further investigate the relation of principal components with hyperplanes in PGD-AT models, we computed the top \( k \) principal components \( P \in \mathbb{R}^{m_0 \times k} \) of the MNIST and Fashion MNIST training datasets and analyzed their similarity with the effective weights \( E_l \in \mathbb{R}^{m_0 \times m_l} \) of the feature network layers in models, given by \( C_l = P^T E_l \in \mathbb{R}^{k \times m_l} \). Results show higher alignment between principal components and hyperplanes in STD-TR models compared to PGD-AT models (see \Cref{fig:fc_pca_components_corr_wrt_weights} and \Cref{fig:appnd_fc_pca_components_corr_wrt_weights_256,fig:appnd_fc_pca_components_corr_wrt_weights_64}). This supports the observation that PGD-AT hyperplanes are positioned to maximize robustness rather than minimize point-to-hyperplane distance, leading to lower similarity with principal components.

\section{Active subnetwork overlap in fully connected robust vs standard models}

Adversaries can alter the output only by changing the active pathways (i.e., NPF). Due to this significance, we measure the overlap in active pathways among samples of the same class and between different classes. The Neural Path Kernel (NPK) $\Psi$ (as per \Cref{eq:npk_definition}) is the gram-matrix of NPFs that measures the overlap of active pathways between pairs of examples. We consider a binary classification task and define two metrics to measure overall NPK overlap between different classes $\Psi^{D}$ and between the same classes $\Psi^{S}$ as defined in \Cref{eq:sameclass_ovrlp,eq:diffclass_ovrlp}. 
\begin{subequations}
    \begin{align}
        \Psi_{\theta}(s,s^{`}) &= <\Phi_{x_{s},\theta},\Phi_{x_{s^{`}},\theta}> \quad s,s^{`} \in [n]\}  \in R^{n,n} \label{eq:npk_definition}\\
        \text{where $\theta$} & \text{ is parameters of the model, $\Phi_{x,\theta} \in R^{P}$ is the NPF} \notag \\ 
        \Psi^{S} &= \sum_{i,j}\Psi_{\theta}(i,j) \quad \forall i,j : y_{true}^{i}=y_{true}^{j} \label{eq:sameclass_ovrlp}\\
        \Psi^{D} &= \sum_{i,j}\Psi_{\theta}(i,j) \quad \forall i,j : y_{true}^{i}\neq y_{true}^{j} \label{eq:diffclass_ovrlp}
    \end{align}
\end{subequations}

\begin{table}
    \small
    \centering
    \begin{tabular}{|m{2.7em} | m{2.3em} | m{2em} | m{2em} || m{2em}| m{2em}|| m{2em}| m{2em}|}
        \toprule
        Dataset & Train Type & $log_{2}$ $\Psi^{D}_{orig}$ & $log_{2}$ $\Psi^{S}_{orig}$ & $log_{2}$ $\Psi^{D}_{adv}$ & $log_{2}$ $\Psi^{S}_{adv}$ & $log_{2}$ $\Psi^{D}_{a,o}$ & $log_{2}$ $\Psi^{S}_{a,o}$\\
        \midrule
        \multirow{2}{2.7em}{MNIST 3vs8} & PGD-AT & \textbf{24.8} & \textbf{26.9} & \textbf{25} & \textbf{26.4} & \textbf{24.8} & \textbf{26.1} \\
        & STD-TR & 24.6 & 27.3 & 27.1 & 28 & 26.9 & 26.2 \\ \hline
        \multirow{2}{2.7em}{MNIST 1vs7} & PGD-AT & \textbf{22.8} & \textbf{27.3} & \textbf{25} & \textbf{26.9} & \textbf{23.9} & \textbf{26.3} \\
        & STD-TR & 20 & 27 & 27.3 & 28 & 25.8 & 26.2 \\ \hline
        \multirow{2}{2.7em}{MNIST 0vs6} & PGD-AT & \textbf{22.9} & \textbf{27.1} & \textbf{24.7} & \textbf{26.4} & \textbf{24} & \textbf{26.3} \\
        & STD-TR & 21.2 & 27.6 & 27.8 & 28.4 & 26.5 & 26.3 \\ \hline
        \multirow{2}{2.7em}{MNIST 1vs5} & PGD-AT & \textbf{22.9} & \textbf{26.9} & \textbf{25} & \textbf{26.4} & \textbf{24.1} & \textbf{26.1} \\
        & STD-TR & 20.42 & 27.5 & 21 & 28.4 & 27.2 & 20.7 \\ \hline
        \multirow{2}{2.7em}{MNIST 3vs9} & PGD-AT & \textbf{24.4} & \textbf{26.8} & \textbf{24.6} & \textbf{26.2} & \textbf{24.4} & \textbf{26} \\ 
        & STD-TR & 23.7 & 27.4 & 26 & 28.3 & 27 & 25 \\ \hline
        \multirow{2}{2.7em}{MNIST 2vs9} & PGD-AT & \textbf{24} & \textbf{27.1} & \textbf{24.6} & \textbf{26.7} & \textbf{24.2} & \textbf{26.5} \\
        & STD-TR & 23.3 & 27.3 & 23.5 & 28.5 & 27 & 23.5\\
        \bottomrule
    \end{tabular}
    \caption{FC-\dlgnmodel-W128-D4 architecture PGD-AT vs STD-TR model subnetwork overlap metrics over original and adversarial examples. The task is binary classification over the MNIST dataset in column 1, and the model has a single output node for classification. PGD-AT rows are highlighted in bold for better readability.}
    \label{tab:npkoverlap_on_pgdat_vs_stdtr}
\end{table}

We obtain these two metrics among adversarial ($\Psi_{adv}$), original samples ($\Psi_{orig}$) and between adversarial and original samples ($\Psi_{adv, org}$) for models trained using PGD-AT and STD-TR on two class datasets (see \Cref{tab:npkoverlap_on_pgdat_vs_stdtr} for MNIST dataset and \Cref{tab:appnd_npkoverlap_on_pgdat_vs_stdtr_famnist} for Fashion MNIST dataset).
Firstly, $\Psi^{D}_{orig} < \Psi^{D}_{adv}$ \& $\Psi^{D}_{orig} < \Psi^{D}_{adv,org}$ for both PGD-AT and STD-TR models. This indicates that adversarial attacks increase active subnetwork overlap between different classes as compared to original samples in an attempt to change the model prediction. Secondly, $\Psi^{D}_{adv,org}$ for PGD-AT is always lesser than $\Psi^{D}_{adv,org}$ for STD-TR models. Also $\Psi^{D}_{adv}$ for PGD-AT is lesser than $\Psi^{D}_{adv}$ for STD-TR models in most cases. These indicate that the active pathways triggered by adversarial examples overlap less with original examples or adversarial examples of another class in the PGD-AT model. Thirdly, $\Psi^{S}_{adv}$ \& $\Psi^{S}_{orig}$ for PGD-AT is always lesser than that in STD-TR. So, the trends so far indicate that the PGD-AT training process learns to map the input to a more diverse path space where overlap among the same class is lesser, and PGD-AT models control subnetwork overlap between different classes during an attack compared to STD-TR models.

\section{Analysis and interpretation of gating patterns in robust vs standard models in convolutional architectures}
\textbf{Notations}
\textit{The following are the notations in convolutional architectures:}
Let $X \in R^{N,1, W, H}$ be the whole training dataset with the size of each sample being $1 \times W \times H$. Let $X_{c} \in R^{N_{c},1, W, H}$ be the training dataset per class with $N_{c}$ being the number of samples of class $c$. Let $L$ be the number of layers, $C_{l}$ be the number of output channels in layer $l$ of feature network (in our experiments for simplicity, we keep $C_{l}$ same across all layers) and $W, H$ be the width, height of output at all feature network layers (since we fix padding=1, kernel size=3, width and height of the output stays same across all layers). Let the output at each feature network layer per class be $F_{l} \in R^{N_{c}, C_{l}, W, H}$. For original examples, let the output combined across all feature network layers be $F^{orig} \in R^{L,N_{c},C_{l},W,H}$ and for adversarial examples let it be $F^{adv} \in R^{L,N_{c},C_{l},W,H}$. "mode" is either adversarial or original examples throughout the paper.

\subsection{Analysis of gating patterns in robust and standard models}
The gates generated in the feature network are the only input representations accessible to the model's value network; hence, their study sheds light on robust models' behaviour. Our goal is to measure the extent of active gate overlap among different class-pairs in \emph{convolutional} \dlgnmodel architectures quantitatively using the idea of intersection-over-union(IOU) and qualitatively by visually inspecting the difference in active gate counts with and without attacks (refer \Cref{tab:appnd_active_gate_count_adv_diff_org_part1,tab:appnd_active_gate_count_adv_diff_org_part2}). The number of active gates per class at each pixel in \( F_{l} \) across all $L$ layers is given by \Cref{eq:active_gate_count_def}.
\begin{subequations}
    \begin{align}
        Gate(x) &=
    \begin{cases}
        1 , & \text{if  } x > 0 \\
        0 , & \text{otherwise}
    \end{cases}\\
    \Lambda^{mode}_{c} &= \sum_{i=1}^{N_{c}}Gate(F^{mode}(X_{c})) , &\in R^{L,C_{l},W,H} \label{eq:active_gate_count_def}    
    \end{align}
\end{subequations}

The procedure to obtain the IOU of active gate count of class $c_1$ and $c_2$ ($IOU_{agc}(c_1,c_2)$) is illustrated in the Appendix.

\begin{table}
    \centering
    \begin{tabular}{|m{1.5em} | m{2em} | m{3.25em}|| m{1.75em} | m{1.75em}| m{1.75em} | m{1.75em}| m{1.75em} |}
        \toprule
        Src Class & Train Type & Quantity & Class 0 & Class 1 & Class 2 & Class 3 & Class 4 \\
        \midrule
        \multirow{4}{1.5em}{0} & \multirow{2}{2em}{PGD-AT} & $IOU^{adv}_{agc}$ & 100 & 70.2 & 83 & 82.7 & 81.8 \\
        & & $IOU^{org}_{agc}$ & 100 & 66.2 & 79.3 & 79.4 & 77.9 \\ \cline{2-8}
        & \multirow{2}{2em}{STD-TR} & $IOU^{adv}_{agc}$ & 100 & 78.1 & 84.7 & 82 & 81 \\
        & & $IOU^{org}_{agc}$ &  100 & 59.7 & 74.7 & 75 & 73.1 \\
        \hline
        \multirow{4}{1.5em}{1} & \multirow{2}{2em}{PGD-AT} & $IOU^{adv}_{agc}$ & 70.2 & 100 & 74.9 & 75.8 & 74.7 \\
        & & $IOU^{org}_{agc}$ & 66.2 & 100 & 71.9 & 74.3 & 71.9 \\ \cline{2-8}
        & \multirow{2}{2em}{STD-TR} & $IOU^{adv}_{agc}$ & 78.1 & 100 & 82.7 & 79.5 & 80 \\
        & & $IOU^{org}_{agc}$ & 59.7 & 100 & 63.7 & 66.5 & 65.6 \\
        \hline
        \multirow{4}{1.5em}{2} & \multirow{2}{2em}{PGD-AT} & $IOU^{adv}_{agc}$ & 83 & 74.9 & 100 & 86.9 & 84.3 \\
        & & $IOU^{org}_{agc}$ & 79.3 & 71.9 & 100 & 84.7 & 80.5 \\ \cline{2-8}
        & \multirow{2}{2em}{STD-TR} & $IOU^{adv}_{agc}$ & 84.7 & 82.7 & 100 & 82.4 & 83.9 \\
        & & $IOU^{org}_{agc}$ & 74.7 & 63.7 & 100 & 80.7 & 74.2 \\
        \hline
        \multirow{4}{1.5em}{3} & \multirow{2}{2em}{PGD-AT} & $IOU^{adv}_{agc}$ & 82.8 & 75.8 & 86.9 & 100 & 82.7 \\
        & & $IOU^{org}_{agc}$ & 79.4 & 74.3 & 84.7 & 100 & 78.6 \\ \cline{2-8}
        & \multirow{2}{2em}{STD-TR} & $IOU^{adv}_{agc}$ & 82 & 79.5 & 82.4 & 100 & 77.5 \\
        & & $IOU^{org}_{agc}$ & 75 & 66.5 & 80.7 & 100 & 73.4 \\
        \hline
        \multirow{4}{1.5em}{4} & \multirow{2}{2em}{PGD-AT} & $IOU^{adv}_{agc}$ & 81.8 & 74.7 & 84.3 & 82.7 & 100 \\
        & & $IOU^{org}_{agc}$ & 77.9 & 71.9 & 80.5 & 78.6 & 100 \\ \cline{2-8}
        & \multirow{2}{2em}{STD-TR} & $IOU^{adv}_{agc}$ & 81 & 80.2 & 83.9 & 77.8 & 100 \\
        & & $IOU^{org}_{agc}$ & 73.1 & 65.6 & 74.2 & 73.4 & 100 \\
        \bottomrule
    \end{tabular}
    \caption{CONV \dlgnmodel-N128-D4 PGD-AT vs STD-TR model IOU of active gate count between class-pairs over adversarial and original examples in MNIST dataset. Only the first four classes are reported here and rest are in \Cref{tab:appnd_iou_agc_class_mnist}}
    \label{tab:iou_agc_class}
\end{table}

We trained a \dlgnmodel with 4 convolutional layers, each having 128 filters (padding 1, stride 1, kernel size 3), followed by an adaptive average pooling layer and a fully connected classification layer. Adversarial training (PGD-AT) was performed on the MNIST dataset with \( \epsilon = 0.3, T = 40, \alpha = 0.005 \), and we measured the Intersection-over-Union (IOU) of active gate overlaps between different class pairs over original ($IOU_{agc}^{org}$) and adversarial ($IOU_{agc}^{adv}$) samples (see \Cref{tab:iou_agc_class} for MNIST and \Cref{tab:appnd_iou_agc_class_fmnist} for Fashion MNIST).
First, $IOU^{org}_{agc}$ for PGD-AT models is consistently higher than for STD-TR models, indicating that gate overlap among classes is initially larger in PGD-AT models. Second, for both PGD-AT and STD-TR, adversarial attacks increase the gate overlap, as $IOU^{adv}_{agc} > IOU^{org}_{agc}$ across all class pairs. Third, the increase in gate overlap ($IOU^{adv}_{agc} - IOU^{org}_{agc}$) is larger in STD-TR models compared to PGD-AT models, demonstrating that minimizing gate overlap among different classes during adversarial attacks is a key feature of PGD-AT models.

\subsection{Interpretation of gating patterns in robust vs standard models}
We aim to further analyze gating patterns by identifying the images that most effectively trigger them. We begin by inverting gating signals in the \dlgnmodel model trained in both PGD-AT and STD-TR modes. Then, we explore more complex gating patterns through inversion. We start by asking: \emph{What is the input image that best simulates the dominant gating signals of an entire class?}. The procedure to obtain such an input image $I$ for class  $c$ is as follows:
\begin{enumerate}
    \item Obtain the active gate count per pixel $\Lambda^{mode}_{c}$ as per \Cref{eq:active_gate_count_def}. Also obtain the inactive gate count per pixel $\eta^{mode}_{c}:\eta^{mode}_{c}(i) = N_c - \Lambda^{mode}_{c}(i)$.
    \item Obtain the dominating gate active-inactive trend per pixel $\rho^{mode}_{c,\lambda}$ as per \Cref{eq:collection_map_vis}.Here $\lambda$ is the threshold which indicates the percentage of gates that has to be active(inactive) among all the class samples to be considered as active(inactive) overall.
    \begin{equation} \label{eq:collection_map_vis}
        \rho^{mode}_{c,\lambda}(i) = 
        \begin{cases}
            1, iff \Lambda^{mode}_{c}(i) > \lambda*N_c\\
            -1, iff \eta^{mode}_{c}(i) > \lambda*N_c
        \end{cases}
    \end{equation}
    \item Let $I$ be the input image under optimization, $F^{mode}$ be the feature maps at the feature network for input $I$ as usual as per our notations. Then, we define a loss function $L(I,\rho^{mode})$ as per \Cref{eq:vis_tanh_loss_def}. This loss function objective is to obtain $I$ such that its feature maps sign at each pixel matches with the dominating gate pattern.
    \begin{equation} \label{eq:vis_tanh_loss_def}
        L(I,\rho^{mode}) = \sum_{i} log(1+e^{-\rho(i)*tanh(F(i))})
    \end{equation}
    \item Now we need to optimize $I$ over the loss function. We explored gaussian blur on gradient and $I$ route but found the results to be satisfactory. However we found the optimization mentioned in \Cref{eq:vis_signed_grad} provides good results.
    \begin{equation} \label{eq:vis_signed_grad}
        I_{t} = I_{t-1} + \alpha sign(\nabla_{I_t}L)
    \end{equation}
    \item Start with $I_{0}=0$ and perform optimization as per \Cref{eq:vis_signed_grad} on the loss function \Cref{eq:vis_tanh_loss_def} for T steps. That is, repeat step 3,4 T times.
\end{enumerate}

\begin{table*}
    \centering
    \begin{tabular}{|m{3.5em} | m{3.5em}| m{3.5em} || m{3.5em} | m{3.5em}||m{3.5em} | m{3.65em} || m{3.5em} | m{3.5em}|}
        \toprule
        Class ($c$) & PGD-AT $I^{org}$ & PGD-AT $I^{adv}$ & STD-TR $I^{org}$ & STD-TR $I^{adv}$ & PGD-AT $I^{ado}$ & PGD-AT $I^{amo}$ & STD-TR $I^{ado}$ & STD-TR $I^{amo}$ \\
        \midrule
        MNIST 5 & \includegraphics[width=\linewidth,keepaspectratio]{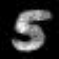} & \includegraphics[width=\linewidth,keepaspectratio]{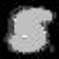} & \includegraphics[width=\linewidth,keepaspectratio]{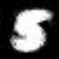} & \includegraphics[width=\linewidth,keepaspectratio]{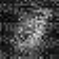} & \includegraphics[width=\linewidth,keepaspectratio]{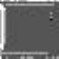} & \includegraphics[width=\linewidth,keepaspectratio]{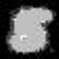} & \includegraphics[width=\linewidth,keepaspectratio]{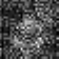} & \includegraphics[width=\linewidth,keepaspectratio]{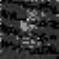}\\
        \hline
        FMNIST Sneaker & 
        \includegraphics[width=\linewidth,keepaspectratio]{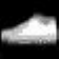} & 
        \includegraphics[width=\linewidth,keepaspectratio]{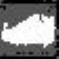} &
        \includegraphics[width=\linewidth,keepaspectratio]{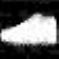} &
        \includegraphics[width=\linewidth,keepaspectratio]{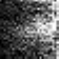} &
        \includegraphics[width=\linewidth,keepaspectratio]{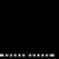} & 
        \includegraphics[width=\linewidth,keepaspectratio]{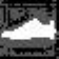} &
        \includegraphics[width=\linewidth,keepaspectratio]{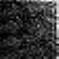} &
        \includegraphics[width=\linewidth,keepaspectratio]{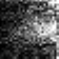}\\
        \bottomrule
    \end{tabular}
    \caption{Image $I$ which triggers dominating gating pattern per class over CONV \dlgnmodel-N128-D4 model obtained on adversarial examples (columns 3,5), obtained on original examples (columns 2,4), obtained on adversarial examples alone but not on original examples (columns 6,8) and obtained both on original examples and adversarial examples(columns 7,9). Visualization loss function is as per \Cref{eq:vis_tanh_loss_def},$\lambda=0.9,\alpha=0.1$,optimization is as per \Cref{eq:vis_signed_grad}. We have reported a few classes for brevity. Detailed results are in \Cref{tab:vis_adv_org_pgdat_stdtr,tab:vis_adv_and_org_pgdat_stdtr}}
    \label{tab:vis_pgdat_stdtr}
\end{table*}

We set $\alpha=0.1$, $T=50$, and $\lambda=0.9$ in our experiments. The visualizations for \dlgnmodel\_N128\_D4 trained on the MNIST, Fashion MNIST dataset are presented in \Cref{tab:vis_pgdat_stdtr}. Dominant gating patterns from original examples capture critical class information, with images inverting these patterns (\(I^{org}\)) clearly resembling their respective classes. The PGD-AT model produces sharper, more distinct class features than the STD-TR model, indicating better utilization of model capacity by PGD-AT. In the STD-TR model, for example, \(I^{org}_5\) can be made to resemble \(I^{org}_8\) with less changes, indicating that one can change input image that triggers dominant gates of class $5$ to the image that triggers dominant gates of class $8$ easily, thereby showing the brittle nature of representations used by STD-TR models. Furthermore, in PGD-AT, \(I^{adv}\) retain class resemblance, while STD-TR's \(I^{adv}\) images are noisy. This indicates PGD-AT prevents adversaries from activating semantically unrelated gating patterns, maintaining class information with slight degradation.
Next, we aim to find the input images (\(I^{ado}\)) that best simulate gating signals dominantly active during adversarial attacks but not in original examples for an entire class and the images (\(I^{amo}\)) that simulate gate patterns active in both adversarial and original examples for an entire class. The visualization process remains the same, except for changes in the computation of \(\rho\). \(I^{ado}_c\) is derived using \(\rho^{ado}_c\) (see \Cref{eq:complex_vis_1}), while \(I^{amo}_c\) uses \(\rho^{amo}_c\) (see \Cref{eq:complex_vis_2}).
\begin{subequations}
    \begin{align}
        \rho^{ado}_{c,\lambda}(i) &= Gate\{\rho^{adv}_{c,\lambda}(i) - Gate(\rho^{org}_{c,\lambda}(i))\}\label{eq:complex_vis_1}\\
        \rho^{amo}_{c,\lambda}(i) &= \rho^{adv}_{c,\lambda}(i) * Gate(\rho^{org}_{c,\lambda}(i)) \label{eq:complex_vis_2}
    \end{align}
\end{subequations}

We report visualized images $I^{ado}_{c},I^{amo}_{c}$ for both \dlgnmodel PGD-AT and STD-TR models as before trained on MNIST, Fashion MNIST dataset in \Cref{tab:vis_pgdat_stdtr}. 
In the PGD-AT model, \(I^{ado}_{c}\) does not produce meaningful inputs, as these patterns are framed images with no resemblance to any class, even using the same visualization method and parameters. This contrasts with \(I^{adv}_{c}\), where adversarial examples resemble class images, indicating that only the dominant active gating patterns from original examples are meaningful in PGD-AT models. In the STD-TR model, both \(I^{adv}_{c}\) and \(I^{ado}_{c}\) appear similar, with \(I^{amo}_{c}\) showing little resemblance to class images, highlighting significant differences between active gates triggered by adversarial and original examples. In PGD-AT, \(I^{amo}_{c}\) shows that adversarial examples trigger a subset of original class gating patterns, maintaining some class resemblance.

\section{Conclusion and future directions}
In this work we utilized \dlgnmodel architectures to thoroughly study the difference in properties exhibited by PGD-AT and STD-TR models. We analyzed fully connected networks, focusing on properties such as hyperplane alignment, path-activity and found that PGD-AT models exhibit larger datapoint separation distances from hyperplanes, active pathways triggered during adversarial attacks in PGD-AT models show less overlap with original examples of other classes and less overlap among original samples of same class suggesting better capacity utilization. We examined convolutional networks to show that PGD-AT models reduce gating overlap among different classes during adversarial attacks. Additionally, we used visualization techniques to understand the dominant gating patterns triggered per class in various scenarios for both STD-TR and PGD-AT models shedding light on the nature of representations used by these models.
We believe that leveraging the results of our analysis to develop novel algorithms that account for the properties examined could effectively enhance robustness. Extending this analysis to larger and more complex models, such as transformers or other deep architectures, could provide further insights into the generalizability of our findings.
While this work focused on PGD-AT, other adversarial training methods could be explored to generalize the understanding of robustness.

\bigskip

\bibliography{aaai25}

\appendix


\crefalias{figure}{appendixfigure}
\crefalias{subfigure}{appendixfigure}
\crefalias{table}{appendixtable}
\crefalias{algorithm}{appendixalg}
\crefalias{equation}{appendixeq}
\crefalias{subequations}{appendixeq}

\section{Appendix} \label{sec:appnd}

\section{Path view of learning in neural networks} \label{sec:appnd_path_view}
\begin{figure*}[t]
    \centering
    \begin{subfigure}[t]{0.19\textwidth}
        \includegraphics[width=\linewidth]{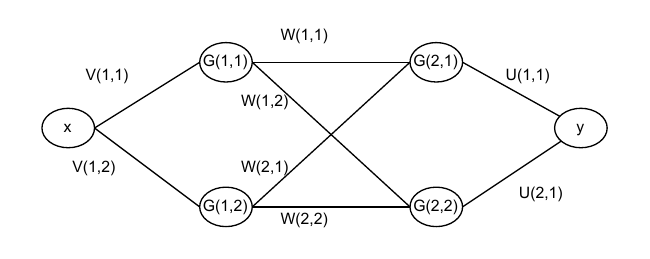}
        \caption{$\hat{y}(x)=$NPF.NPV}
    \end{subfigure} \hfill
    \begin{subfigure}[t]{0.19\textwidth}
        \includegraphics[width=\linewidth]{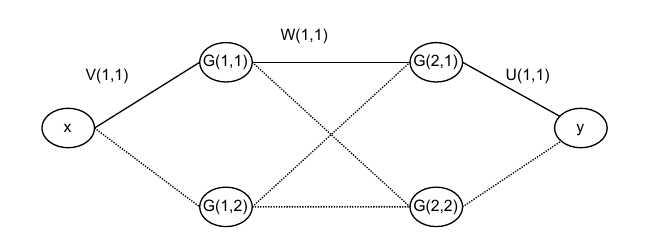}
        \caption{NPV1=V(1,1)W(1,1)U(1,1) NPF1=x*G(1,1)G(2,1)}
    \end{subfigure} \hfill
    \begin{subfigure}[t]{0.19\textwidth}
        \includegraphics[width=\linewidth]{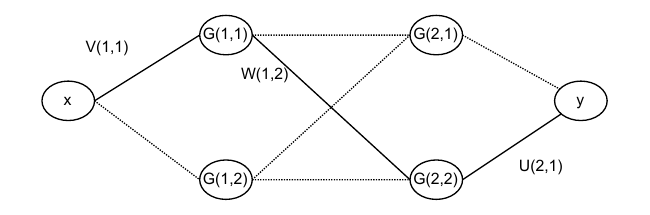}
        \caption{NPV2=V(1,1)W(1,2)U(2,1) NPF2=x*G(1,1)G(2,2)}
    \end{subfigure} \hfill
    \begin{subfigure}[t]{0.19\textwidth}
        \includegraphics[width=\linewidth]{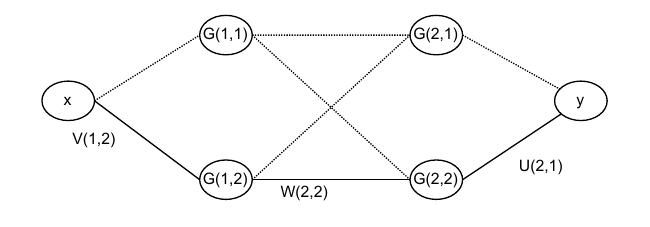}
        \caption{NPV4=V(1,2)W(2,2)U(2,1) NPF4=x*G(1,2)G(2,2)}
    \end{subfigure} \hfill
    \begin{subfigure}[t]{0.19\textwidth}
        \includegraphics[width=\linewidth]{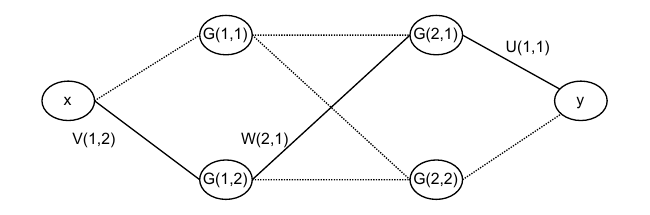}
        \caption{NPV3=V(1,2)W(2,1)U(1,1) NPF3=x*G(1,2)G(2,1)}
    \end{subfigure} \hfill
    \caption{Expressing network output as combination of paths. This gives the path-view of learning in neural networks. Output at a output node is the dot product of NPF and NPV vectors(which are in path space). }
    \label{fig:path_view_neural_networks}
\end{figure*}

An input to a neural network (with RELU activations) leads to a certain subnetwork being active. The input can be viewed as being mapped to path space wherein the path space representation is given by the neural path feature (NPFs), and the weight of each path is given by neural path value (NPVs). NPF of a path is given by the product of gates along a path and NPV of a path is given by the product of weights along a path. \Cref{fig:path_view_neural_networks} illustrates this idea of path-view along with concept of NPF,NPV. The overall network output can be obtained as dot product of the NPFs and NPVs as demonstrated in \Cref{eq:npf_dot_npv_proof}. 
\begin{subequations} \label{eq:npf_dot_npv_proof}
    \begin{align}
        NPF.NPV &= V(1,1)W(1,1)U(1,1)x*G(1,1)G(2,1) \notag \\
                &\quad + V(1,2)W(2,2)U(2,1)x*G(1,2)G(2,2) \notag \\
                &\quad + V(1,1)W(1,2)U(2,1)x*G(1,1)G(2,2) \notag \\
                &\quad + V(1,2)W(2,1)U(1,1)x*G(1,2)G(2,1) \notag \\
        NPF.NPV &= U(1,1)G(2,1)\big[x*V(1,1)G(1,1)W(1,1) \notag \\
                &\quad + x*V(1,2)G(1,2)W(2,1)\big] \notag \\
                &\quad + U(2,1)G(2,2)\big[x*V(1,1)G(1,1)W(1,2) \notag \\
                &\quad + x*V(1,2)G(1,2)W(2,1)\big] \notag \\
        NPF.NPV &= U(1,1)RELU(2,1) + U(2,1)RELU(2,2) \notag \\
        NPF.NPV &= \hat{y}
    \end{align}
\end{subequations}

\begin{algorithm}
    \caption{PGD adversarial training for $M$ epochs, given some radius $\epsilon$, adversarial step size $\alpha$, $T$ PGD steps and a dataset of size $N$ for a network $F_\theta$}
    \label{alg:pgd_at}
        \begin{algorithmic}
        \FOR {$j=1\dots M$}
        \FOR {$i=1\dots N$}
        \STATE \textit{// Perform PGD adversarial attack}
        \STATE $\delta = U(-\epsilon,\epsilon)$
        \FOR {$t=1\dots T$}
        \STATE $\delta = \delta + \alpha \cdot \sign(\nabla_\delta L(F_\theta(x_i + \delta),y_i^{true}))$
        \STATE $\delta = \max(\min(\delta, \epsilon), -\epsilon)$
        \ENDFOR
        \STATE $\theta = \theta - \nabla_\theta L(F_\theta(x_i + \delta), y_i)$ \textit{// Update model weights with some optimizer, e.g. SGD}
        \ENDFOR
        \ENDFOR
        \end{algorithmic}
\end{algorithm}


PGD-40 and clean accuracies over MNIST and Fashion MNIST dataset using various architectures are reported at \Cref{tab:appnd_pgdat_stdtr_accuracies}.

\begin{table}
    \centering
    \begin{tabular}{|m{2.6em}| m{4.5em} | m{3em} || m{6em}| m{3em}|}
        \toprule
        Dataset & Architecture & Training Type & PGD-40 Test Acc. \((@\epsilon=0.3,@\epsilon=0.2)\) & Clean Test Acc. \\
        \midrule
        \multirow{4}{2em}{MNIST} & FC-\dlgnmodel-W128-D4 & PGD-AT & (49.8\%,54.5\%) & 66.4\%\\ \cline{2-5}
        & FC-\dlgnmodel-W128-D4 & STD-TR & (1.6\%,2.7\%) & 97.7\%\\ \cline{2-5}
        & CONV-\dlgnmodel-N128-D4 & PGD-AT & (78.9\%,88.9\%) & 97.5\%\\ \cline{2-5}
        & CONV-\dlgnmodel-N128-D4 & STD-TR & (0.05\%,0.06\%) & 98.4\%\\
        \hline
        \multirow{4}{2em}{Fashion MNIST} & FC-\dlgnmodel-W128-D4 & PGD-AT & (40.2\%,48.3\%) & 62.6\%\\ \cline{2-5}
        & FC-\dlgnmodel-W128-D4 & STD-TR & (3.7\%,5.1\%) & 88.6\%\\ \cline{2-5}
        & CONV-\dlgnmodel-N128-D4 & PGD-AT & (49.8\%,88.9\%) & 67.8\%\\ \cline{2-5}
        & CONV-\dlgnmodel-N128-D4 & STD-TR & (0\%,0\%) & 88.9\%\\
        \bottomrule
    \end{tabular}
    \caption{PGD-AT vs STD-TR model PGD accuracies and clean accuracies}
    \label{tab:appnd_pgdat_stdtr_accuracies}
\end{table}

\subsection{More analysis of hyperplanes in feature network of PGD-AT and STD-TR models} \label{sec:appnd_hp_analysis}
\begin{figure}
  \begin{subfigure}[b]{0.5\textwidth}
  \includegraphics[width=\linewidth]{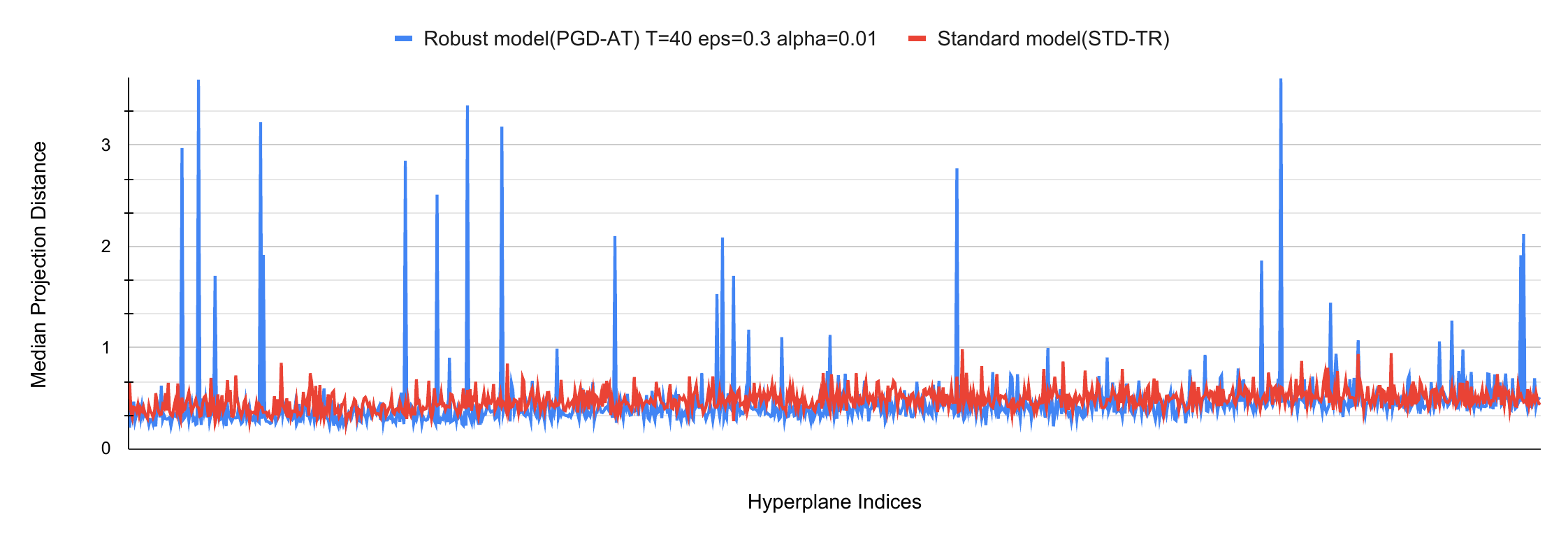}
  \end{subfigure}\\ 
    \begin{subfigure}[b]{0.5\textwidth}
    \includegraphics[width=\linewidth]{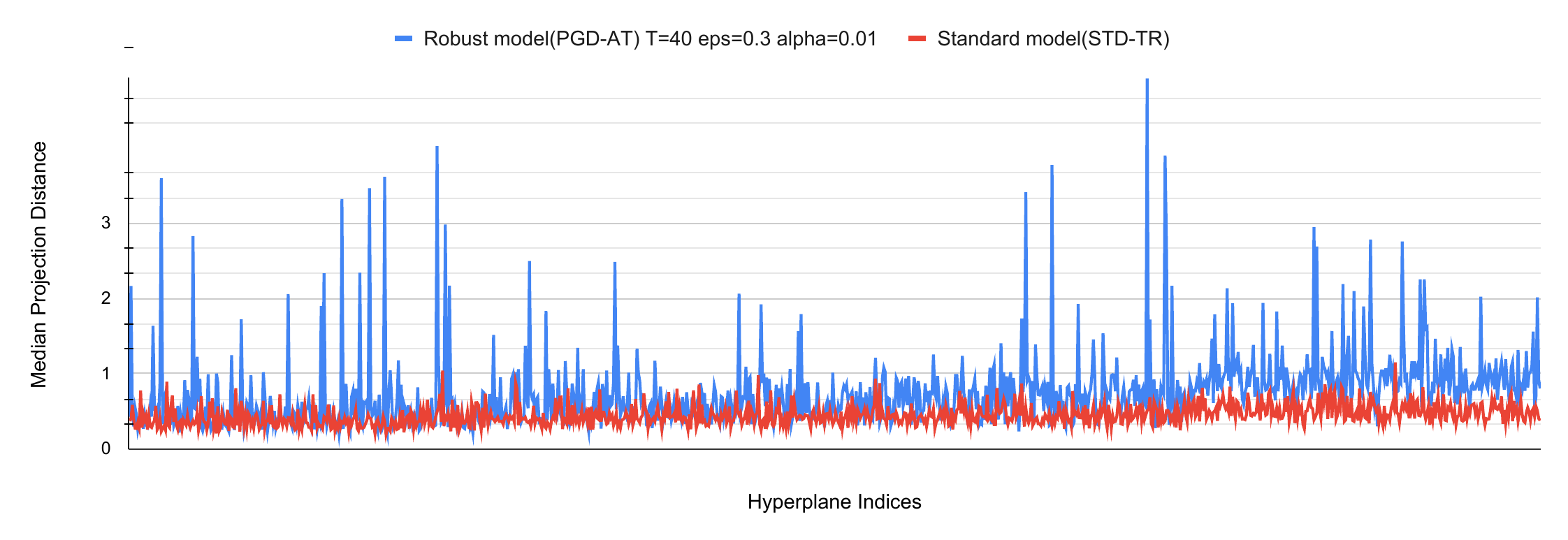}
    \end{subfigure}
\caption{PGD-AT vs STD-TR FC-\dlgnmodel-W256-D4 median projection distance. The top image is on MNIST, and the bottom image is on the Fashion MNIST dataset. The Y-axis denotes the median projection distance of data points at node/hyperplane indices on the X-axis.}
\label{fig:appnd_fc_dlgn_median_distance_from_HP_256}
\end{figure}

\begin{figure}
\begin{subfigure}[b]{0.5\textwidth}
  \includegraphics[width=\linewidth]{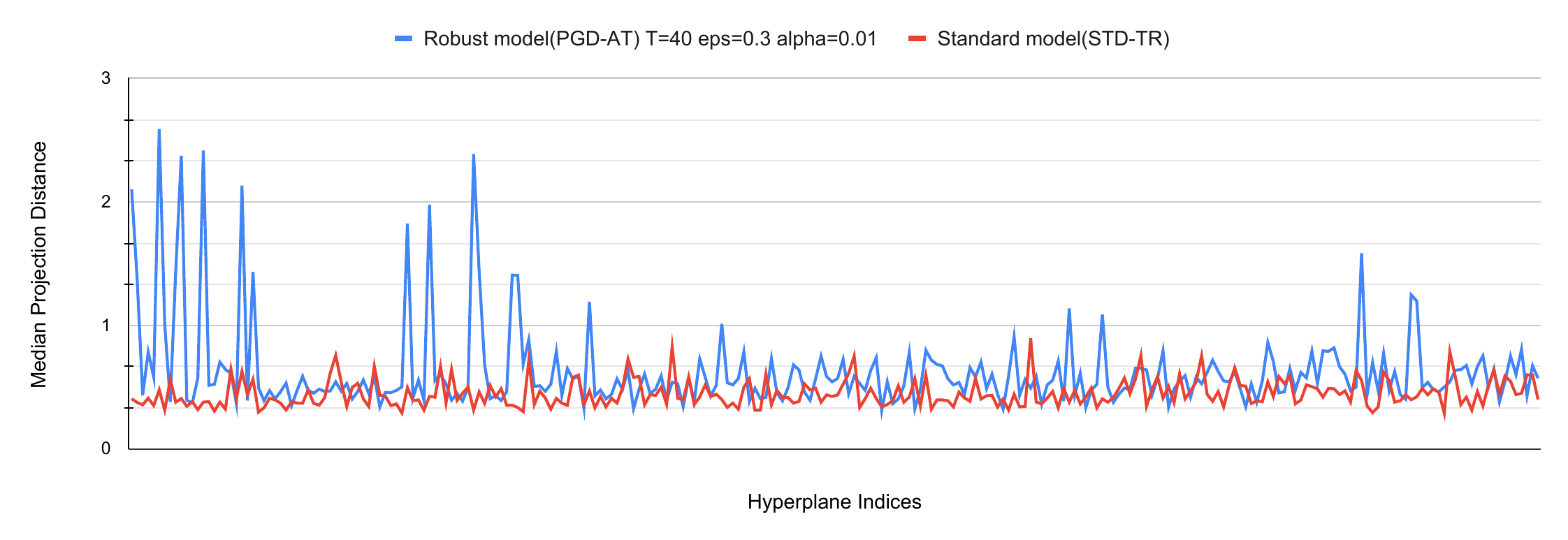}
  \end{subfigure}\\
  \begin{subfigure}[b]{0.5\textwidth}
    \includegraphics[width=\linewidth]{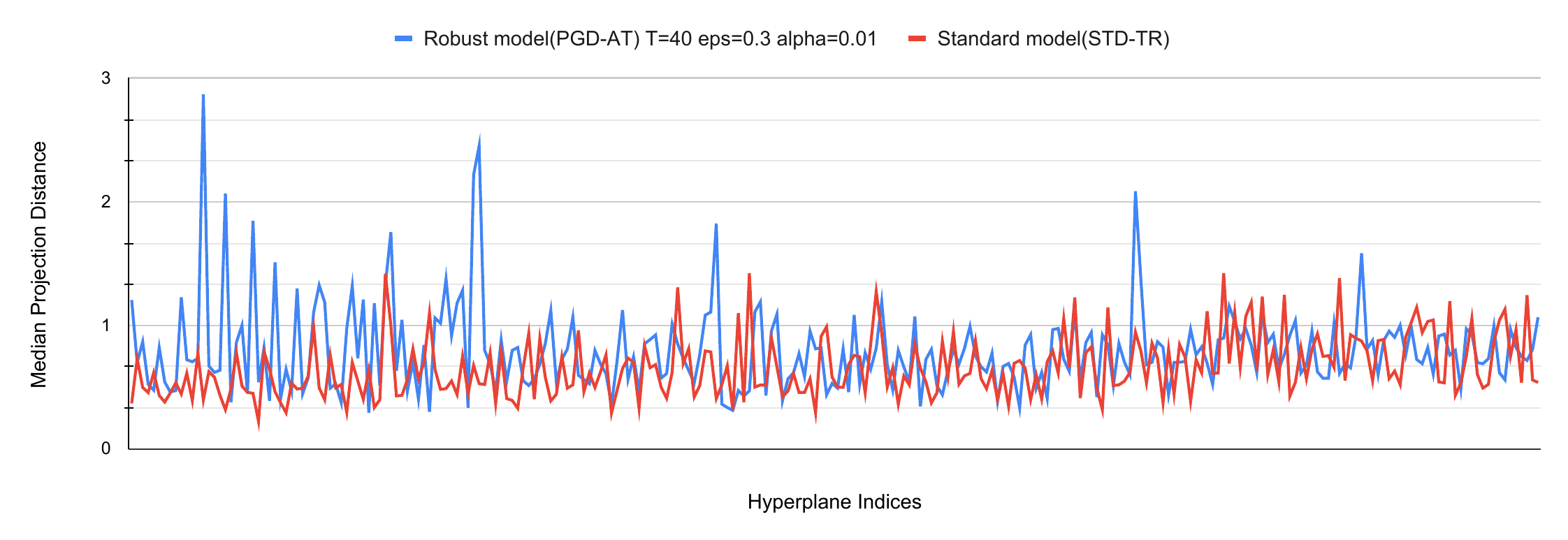}
    \end{subfigure}
\caption{PGD-AT vs STD-TR FC-\dlgnmodel-W64-D4 median projection distance. The left image is on MNIST, and the right image is on the Fashion MNIST dataset. The Y-axis denotes the median projection distance of data points at node/hyperplane indices on the X-axis.}
\label{fig:appnd_fc_dlgn_median_distance_from_HP_64}
\end{figure}

The median projection distance at each hyperplane in PGD-AT and STD-TR models of \dlgnmodel with width 256 (see \Cref{fig:appnd_fc_dlgn_median_distance_from_HP_256}) and 64 (see \Cref{fig:appnd_fc_dlgn_median_distance_from_HP_64}) also clearly shows that median distances increase in robust models.

\begin{figure*}
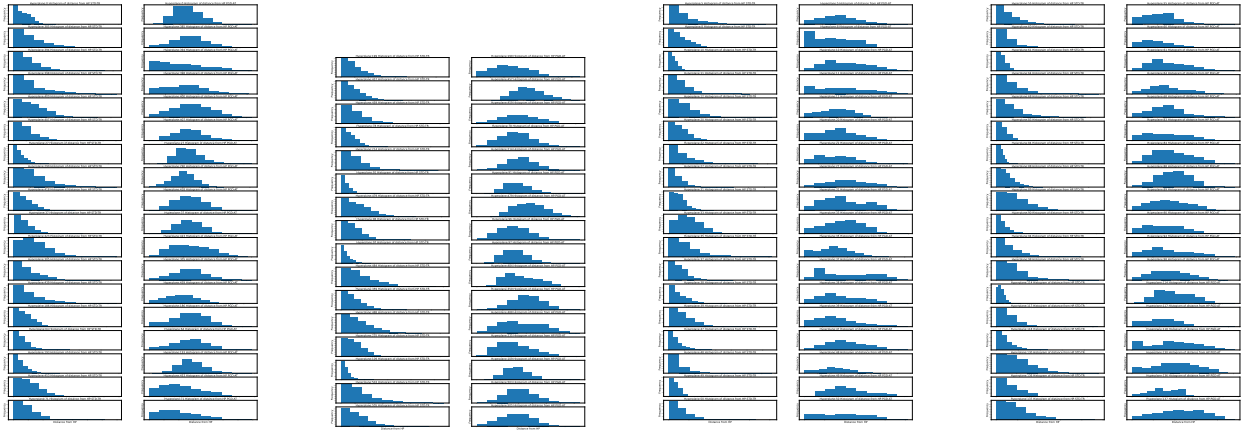

  \begin{subfigure}[b]{0.24\textwidth}
  \includegraphics[width=\linewidth]{appendix/images/distance_from_HP/fc_dlgn_sig_gate_wise_histogram_pgd_vs_std_sigdist_0.5_1.pdf}
  \end{subfigure}
  \begin{subfigure}[b]{0.24\textwidth}
  \includegraphics[width=\linewidth]{appendix/images/distance_from_HP/fc_dlgn_sig_gate_wise_histogram_pgd_vs_std_sigdist_0.5_2.pdf}
  \end{subfigure}
  \begin{subfigure}[b]{0.24\textwidth}
  \includegraphics[width=\linewidth]{appendix/images/distance_from_HP/fc_dlgn_sig_gate_wise_histogram_pgd_vs_std_sigdist_0.5_1_famnist.pdf}
  \end{subfigure}
  \begin{subfigure}[b]{0.24\textwidth}
  \includegraphics[width=\linewidth]{appendix/images/distance_from_HP/fc_dlgn_sig_gate_wise_histogram_pgd_vs_std_sigdist_0.5_2_famnist.pdf}
  \end{subfigure}
\caption{Projection distance distribution at hyperplanes whose medians differ significantly(by 0.5) between standard and robust \dlgnmodel models. Each row in each image denotes a hyperplane, with the Y-axis indicating the frequency of occurrence and the X-axis being the distance from that row's hyperplane. Columns 1,3 are for the STD-TR model, and columns 2,4 are for the PGD-AT model. Both X \& Y axis is shared per row. First-row images correspond to the MNIST dataset, and the second-row images correspond to the Fashion MNIST dataset.}
\label{fig:appnd_projection_distance_distribution}.
\end{figure*}

The projection distance histogram at hyperplanes, which shows significant
differences in median projection distance between standard and robust models (see \Cref{fig:appnd_projection_distance_distribution}), also shows that the projection distance of datapoints is shifted to larger distances in PGD-AT than STD-TR models. 

\subsection{Hyperplane analysis in synthetic XOR dataset} \label{subsec:appnd_hp_analysis_xor}
\begin{figure}[H]
    \centering
    \includegraphics[width=0.3\textwidth]{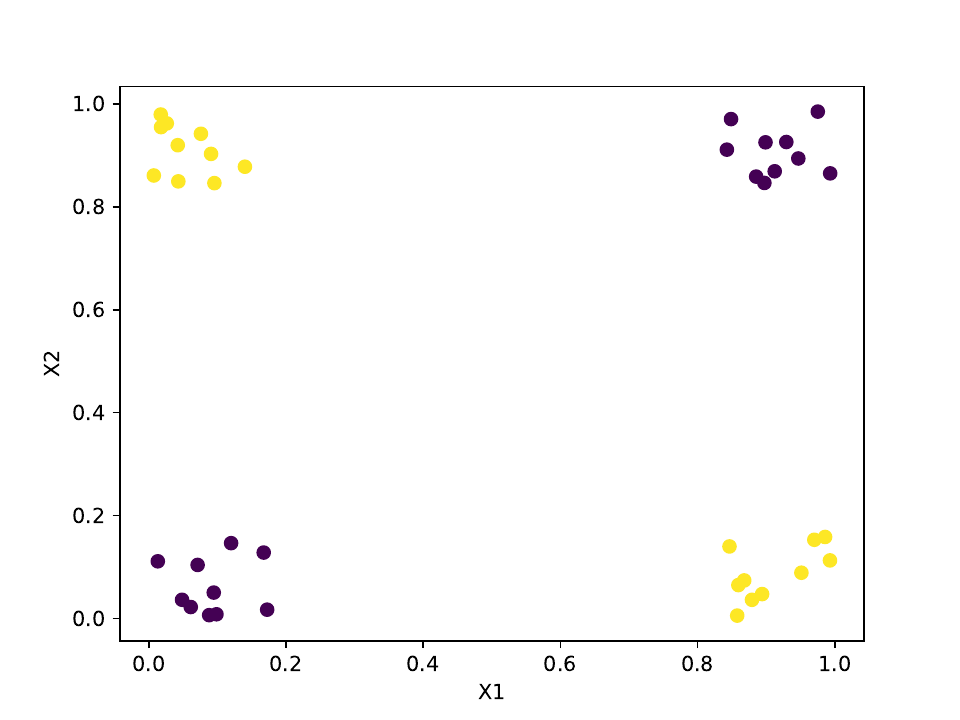} 
    \caption{2D XOR dataset with gap from x=0.5,y=0.5 being 0.32 to facilitate PGD-AT with eps < 0.32.}
    \label{fig:xor_dataset_train}
\end{figure}

\begin{figure}[H]
\begin{subfigure}{0.23\textwidth}
    \includegraphics[width=\textwidth]{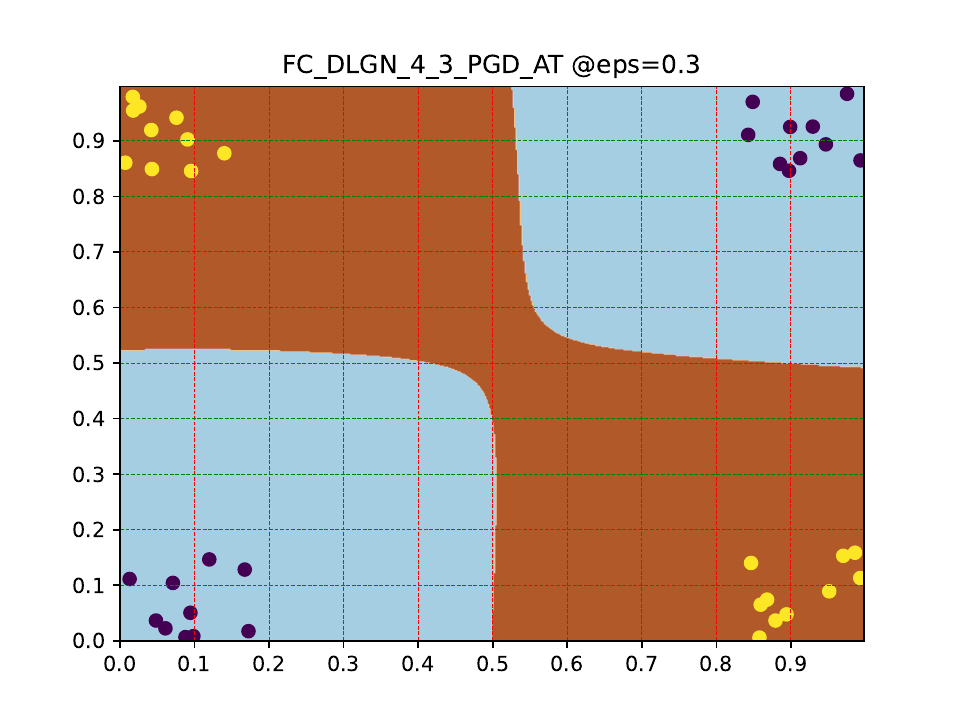}
    \caption{PGD-AT decision boundary.}
    \label{fig:xor_decision_boundary_pgdat}
\end{subfigure}
\begin{subfigure}{0.23\textwidth}
    \includegraphics[width=\textwidth]{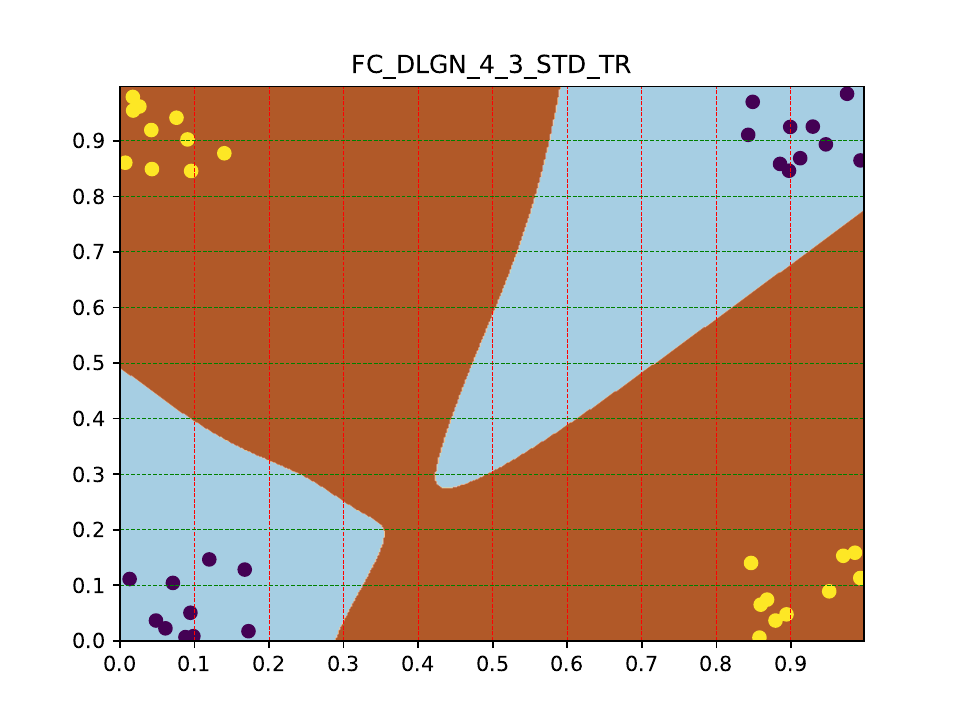}
    \caption{STD-TR decision boundary.}
    \label{fig:xor_decision_boundary_stdtr}
\end{subfigure}
\end{figure}

The synthetic XOR 2D dataset constructed with a gap from x=0.5 and y=0.5 axis is shown in \Cref{fig:xor_dataset_train}. The decision boundaries of PGD-AT models (see \Cref{fig:xor_decision_boundary_pgdat}) are closer to optimal compared to STD-TR (see \Cref{fig:xor_decision_boundary_stdtr}), ensuring that adversarial examples within \( L_\infty \) bounds (\( \epsilon = 0.3 \)) are correctly classified only by PGD-AT.


\subsection{PCA ANALYSIS IN ROBUST AND STANDARD MODELS} \label{subsec:appnd_pca_analysis}

\begin{figure}[H]
  \begin{subfigure}[b]{0.5\textwidth}
  \includegraphics[width=\linewidth]{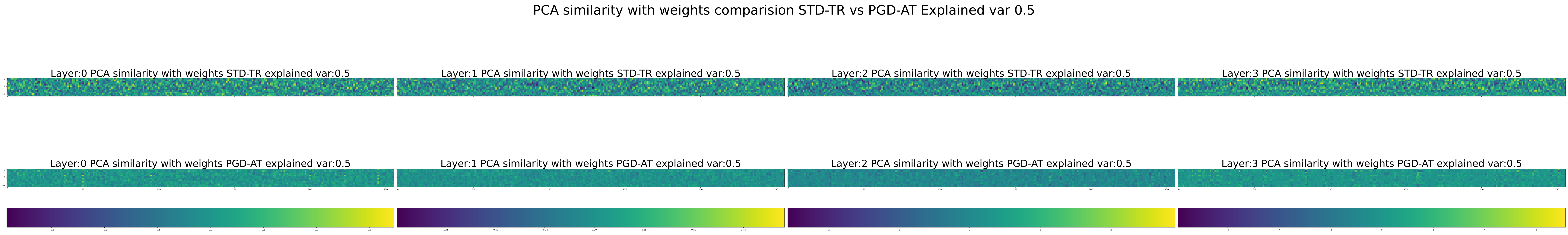}
  \caption{MNIST dataset with explained variance $0.5$, $12$ components.}
  \end{subfigure}
  \begin{subfigure}[b]{0.5\textwidth}
  \includegraphics[width=\linewidth]{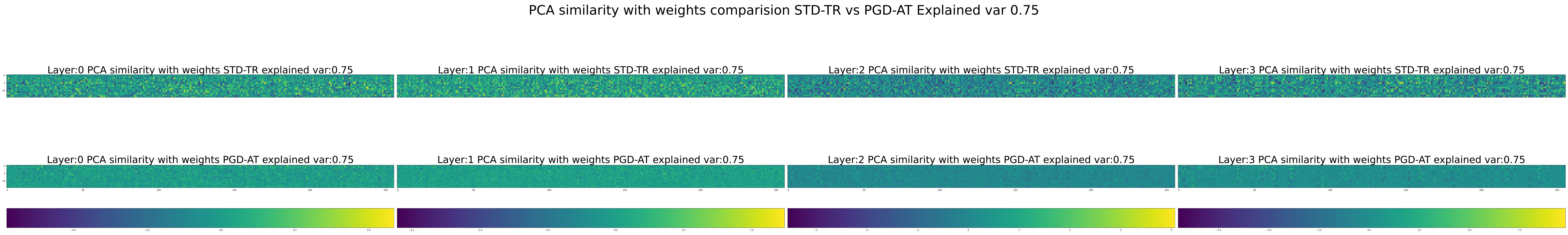}
  \caption{Fashion MNIST dataset with explained variance $0.75$, $15$ components.}
  \end{subfigure}\\
\caption{Effective weights with top PCA components in PGD-AT(bottom row) and STD-TR(top row) using FC-\dlgnmodel-W256-D4 architecture.}
\label{fig:appnd_fc_pca_components_corr_wrt_weights_256}
\end{figure}

\begin{figure}[H]
  \begin{subfigure}[b]{0.5\textwidth}
  \includegraphics[width=\linewidth]{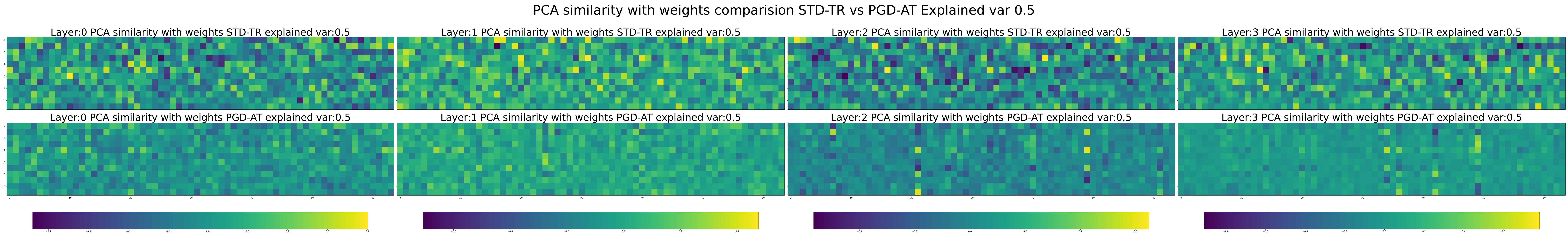}
  \caption{MNIST dataset with explained variance $0.5$, $12$ components.}
  \end{subfigure}
  \begin{subfigure}[b]{0.5\textwidth}
  \includegraphics[width=\linewidth]{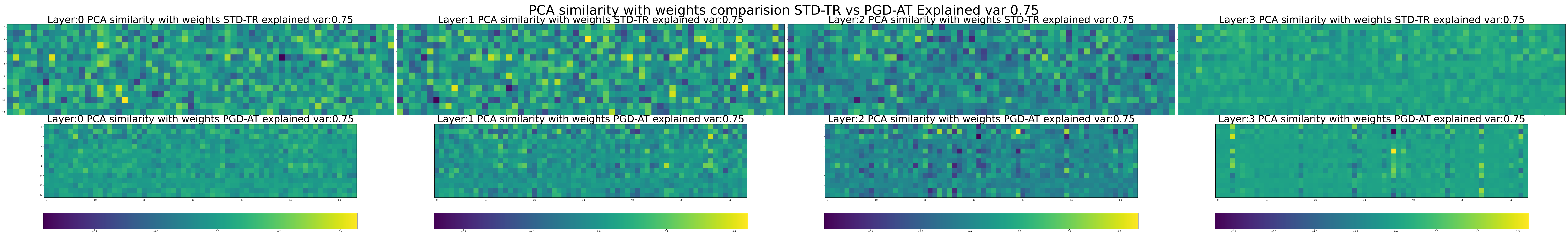}
  \caption{Fashion MNIST dataset with explained variance $0.75$, $15$ components.}
  \end{subfigure}
\caption{Effective weights with top PCA components in PGD-AT(bottom row) and STD-TR(top row) using FC-\dlgnmodel-W64-D4 architecture.}
\label{fig:appnd_fc_pca_components_corr_wrt_weights_64}
\end{figure}

We report similarity of principal components with hyperplanes of feature network of DLGN with width 256 in \Cref{fig:appnd_fc_pca_components_corr_wrt_weights_256} and width 64 in \Cref{fig:appnd_fc_pca_components_corr_wrt_weights_64} respectively.

\subsection{More results in Active subnetwork overlap in PGD-AT vs STD-TR models} \label{subsec:appnd_subnet_ovrlp}

The subnetwork overlap metrics for FC-\dlgnmodel\_W128\_D4 architecture trained over the Fashion MNIST dataset is shown in \Cref{tab:appnd_npkoverlap_on_pgdat_vs_stdtr_famnist}.

\begin{table*}
    \centering
    \begin{tabular}{|m{3.5em} | m{3.75em}| m{3em} | m{3em} | m{2.5em} | m{2.5em} || m{2.5em}| m{2.5em}|| m{2.5em}| m{2.5em}|}
        \toprule
        Dataset & Train Type & PGD-40 Acc. & Clean Acc.& $log_{2}$ $\Psi^{D}_{orig}$ & $log_{2}$ $\Psi^{S}_{orig}$ & $log_{2}$ $\Psi^{D}_{adv}$ & $log_{2}$ $\Psi^{S}_{adv}$ & $log_{2}$ $\Psi^{D}_{adv,or}$ & $log_{2}$ $\Psi^{S}_{adv,or}$\\
        \midrule
        \multirow{2}{3.5em}{FaMNIST 1vs9} & PGD-AT & 93.90\%  & 99.70\% & 22.78 & 30.45 & \textbf{28.60} & \textbf{31.01} & \textbf{26.43} & \textbf{29.81} \\
        & STD-TR  & 0.00\% & 100.00\%  & 28.37 & 31.31  & 31.25 & 32.41 & 31.26 & 30.01 \\ \cline{1-10}
        \multirow{2}{3.5em}{FaMNIST 3vs8} & PGD-AT  & 76.75\% & 90.45\% & \textbf{25.92} & \textbf{29.60} & \textbf{28.58} & \textbf{29.29} & \textbf{27.53} & \textbf{28.74} \\
        & STD-TR & 4.65\% & 99.30\% & 25.88 & 30.39  & 30.41 & 31.68  & 29.04 & 30.01 \\ \cline{1-10}
        \multirow{2}{3.5em}{FaMNIST 7vs9} & PGD-AT & 80.75\% & 87.30\% & \textbf{26.28} & \textbf{29.05} & \textbf{28.29} & \textbf{28.98} & \textbf{27.38} & \textbf{28.54} \\
        & STD-TR & 0.00\% & 97.00\%  & 26.58  & 29.47  & 31.27 & 31.67  & 29.48 & 29.61 \\ \cline{1-10}
        \multirow{2}{3.5em}{FaMNIST 0vs2} & PGD-AT & 74.25\% & 90.10\% & \textbf{26.68} & \textbf{30.04} & \textbf{27.61} & \textbf{29.00}  & \textbf{27.06} & \textbf{29.01} \\
        & STD-TR & 0.00\% & 97.10\% & 28.74  & 30.79  & 30.10  & 32.15 & 30.73 & 29.17 \\ \cline{1-10}
        \multirow{2}{3.5em}{FaMNIST 4vs5} & PGD-AT & 92.75\% & 98.90\% & \textbf{22.70} & \textbf{29.82} & \textbf{29.87} & \textbf{30.24} & \textbf{28.87} & \textbf{29.09} \\
        & STD-TR & 23.80\% & 99.00\% & 27.88 & 31.16 & 31.31 & 31.91 & 30.55 & 30.08\\ \cline{1-10}
        \multirow{2}{3.5em}{FaMNIST 6vs7} & PGD-AT & 89.00\% & 98.40\% & \textbf{23.36} & \textbf{29.98} & \textbf{28.47} & \textbf{31.02} & \textbf{31.32} & \textbf{30.30} \\
        & STD-TR  & 23.80\% & 100.00\% & 27.35 & 31.50 & 23.50 & 29.21  & 27.44 & 28.73\\  \bottomrule
    \end{tabular}
    \caption{FC-\dlgnmodel-W128-D4 architecture PGD-AT vs STD-TR model path overlaps metrics over original and adversarial examples. The task is binary classification over the Fashion MNIST dataset in column 2, and the model has a single output node for classification. PGD-AT rows are highlighted in bold for better readability.}
    \label{tab:appnd_npkoverlap_on_pgdat_vs_stdtr_famnist}
\end{table*}

\section{Qualitative analysis of gating patterns in PGD-AT and STD-TR models} \label{app:appnd_qualitative_gate}

We qualitatively inspect the difference in active gate counts with and without attacks using $\Lambda^{adv\_diff\_org}_{c}$ in \Cref{eq:active_gate_count_adv_diff_org} that measures the difference in active gate count for adversarial and original examples and is plotted per class for both PGD-AT and STD-TR models as an image of size $L*C_{l}, W, H$ in \Cref{tab:appnd_active_gate_count_adv_diff_org_part1} and \Cref{tab:appnd_active_gate_count_adv_diff_org_part2}.

\begin{subequations}
    \begin{align}
        \Lambda^{mode}_{c} &= \sum_{i=1}^{N_{c}}Gate(F^{mode}(X_{c})) , &\in R^{L,C_{l},W,H} \label{eq:appnd_active_gate_count_def} \\
        \intertext{where $mode$ is either original examples or adversarial examples}
        \Lambda^{\mathit{adv\_diff\_org}}_{c}(i) &= \Lambda^{adv}_{c}(i) - \Lambda^{org}_{c}(i) , & \forall \: i \in R^{L,C_{l},W,H} \label{eq:active_gate_count_adv_diff_org}
    \end{align}
\end{subequations}

\begin{table*}
    \centering
    \begin{tabular}{|m{2em} || m{16em}| m{16em}|}
        \toprule
        Class ($c$) & PGD-AT $\Lambda^{\mathit{adv\_diff\_org}}_{c}$ & STD-TR $\Lambda^{\mathit{adv\_diff\_org}}_{c}$ \\
        \midrule
        0 & \includegraphics[width=16em,keepaspectratio]{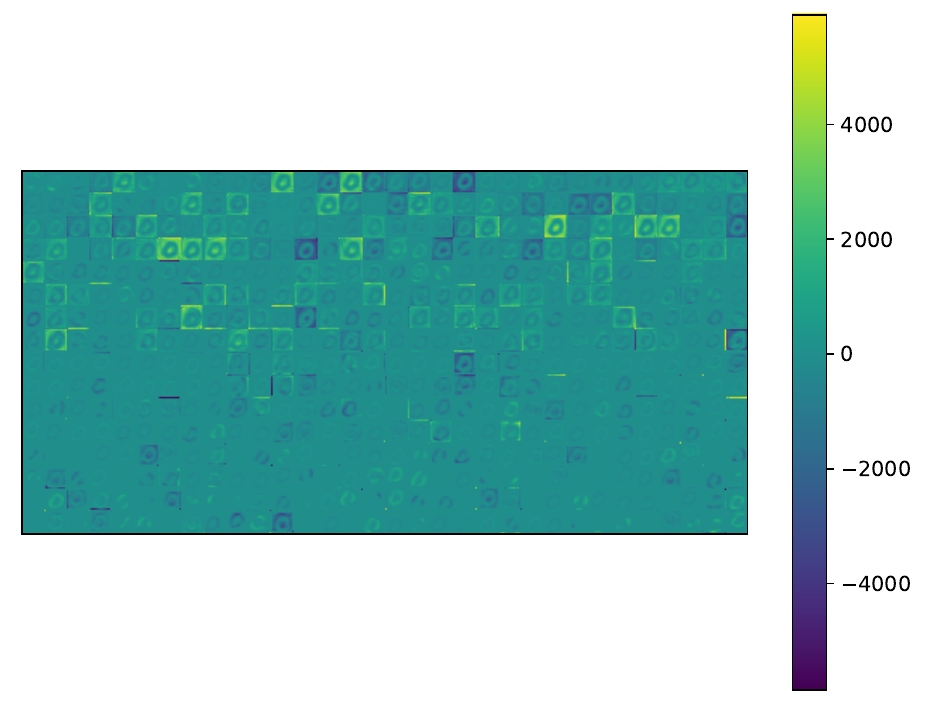} & \includegraphics[width=16em,keepaspectratio]{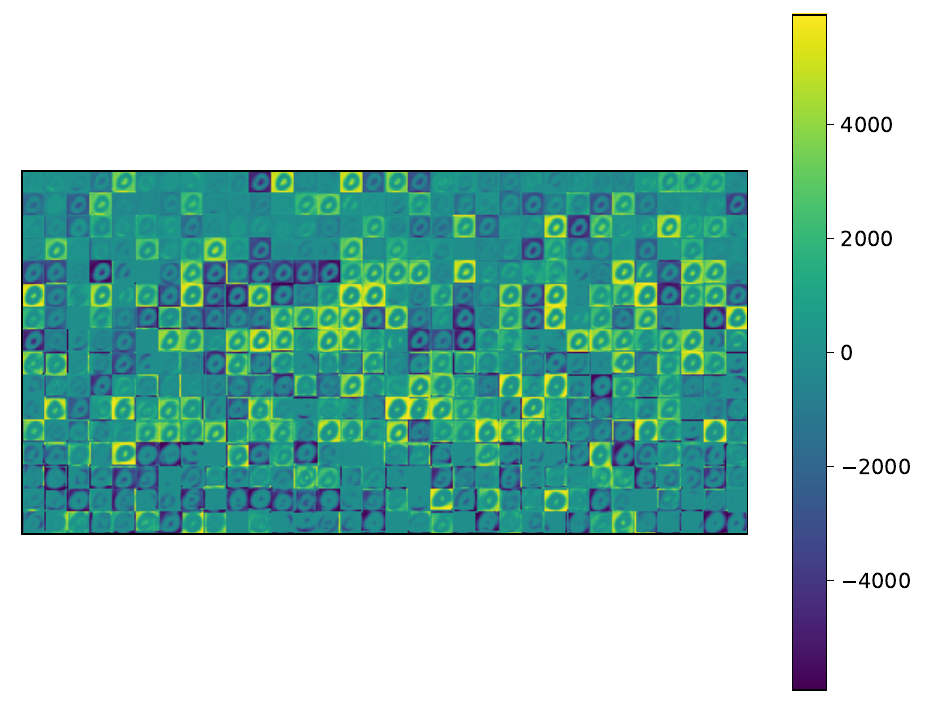}\\
        \hline
        1 & \includegraphics[width=16em,keepaspectratio]{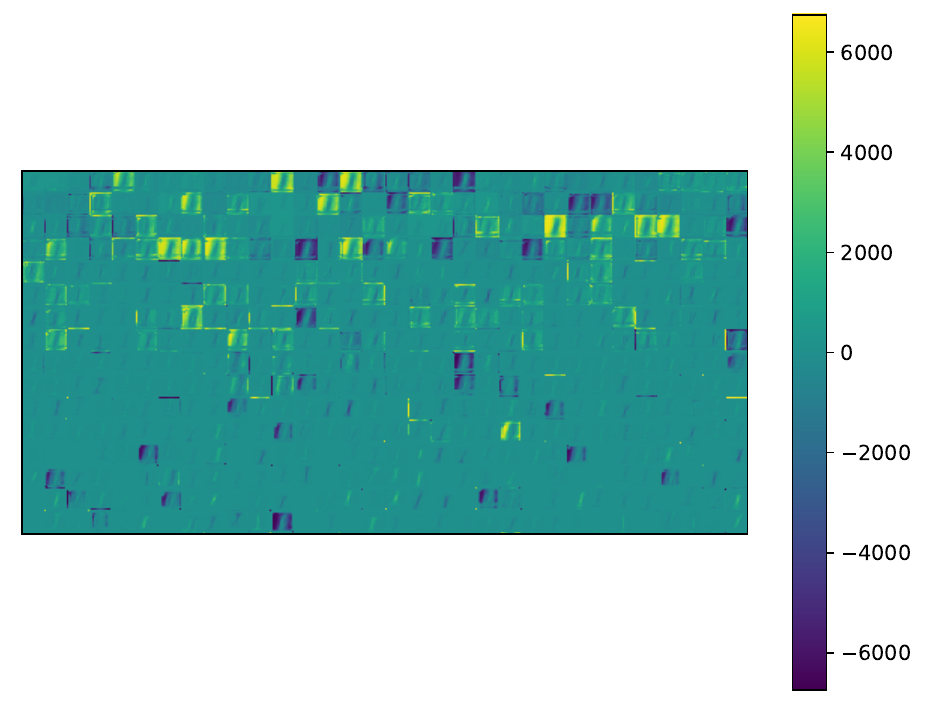} & \includegraphics[width=16em,keepaspectratio]{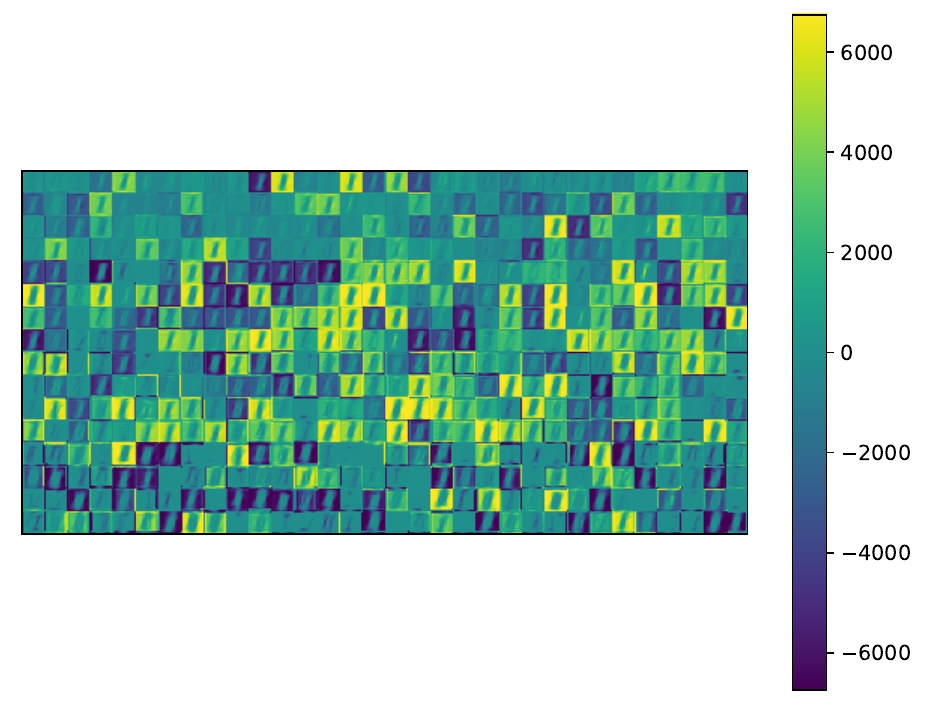}\\
        \hline
        2 & \includegraphics[width=16em,keepaspectratio]{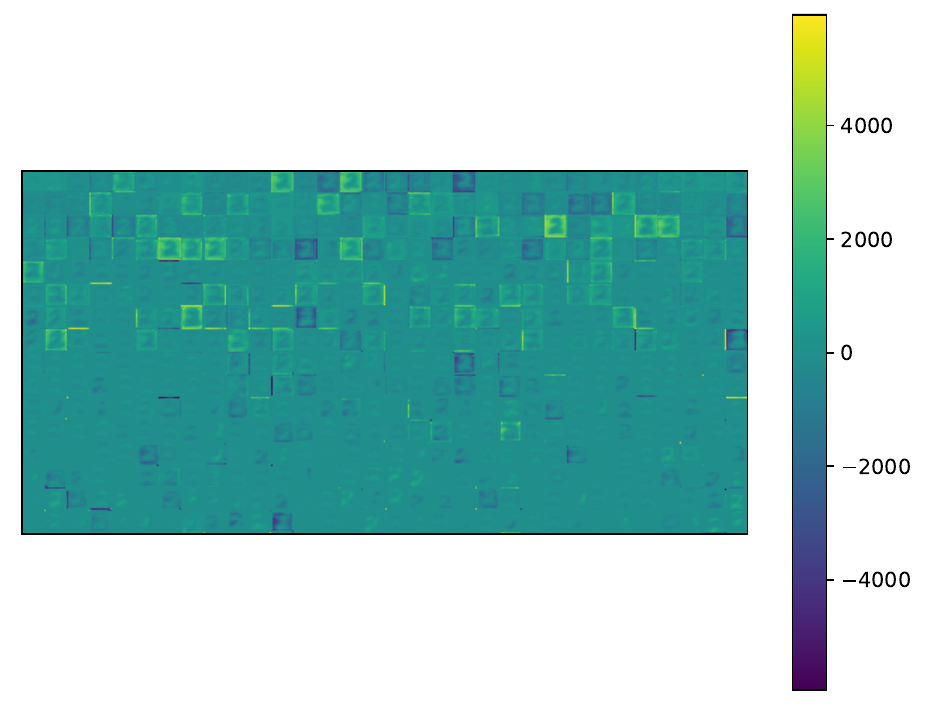} & \includegraphics[width=16em,keepaspectratio]{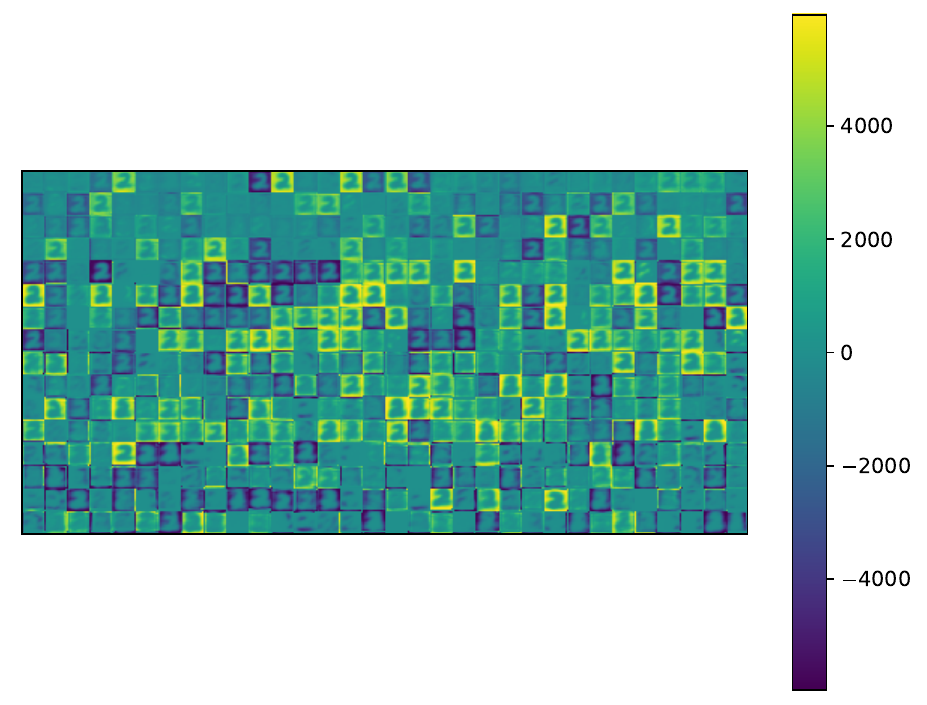}\\
        \hline
        3 & \includegraphics[width=16em,keepaspectratio]{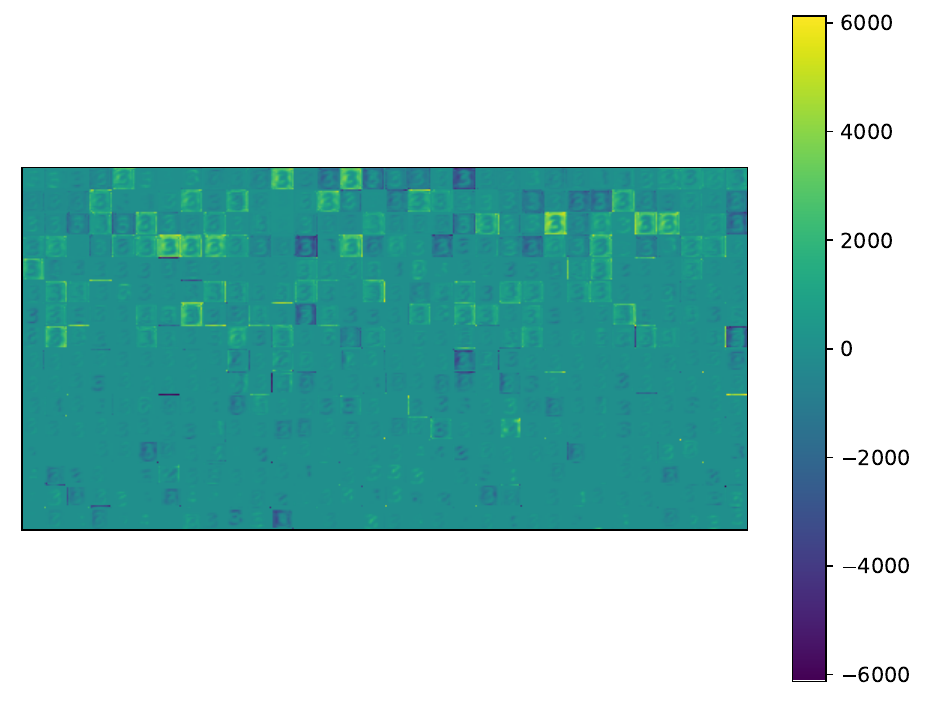} & \includegraphics[width=16em,keepaspectratio]{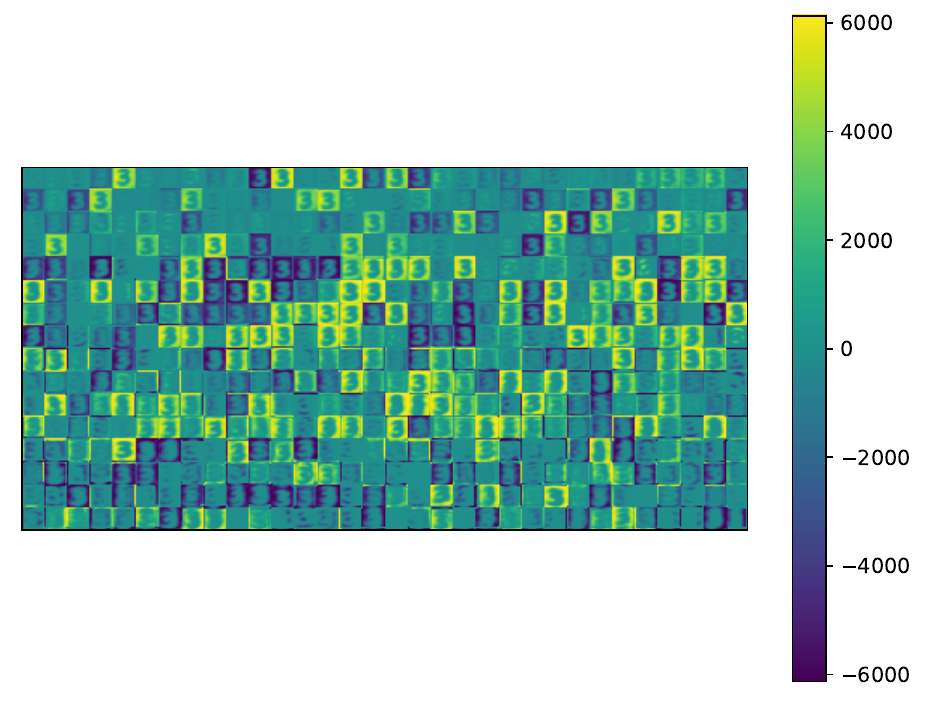}\\
        \bottomrule
    \end{tabular}
    \caption{$\Lambda^{\mathit{adv\_diff\_org}}_{c}$ for PGD-AT and STD-TR models with CONV\_N128\_D4 \dlgnmodel architecture on MNIST dataset. In each cell of the image, every four rows represent a layer's $\Lambda^{\mathit{adv\_diff\_org}}_{l,c}$}
    \label{tab:appnd_active_gate_count_adv_diff_org_part1}
\end{table*}

\begin{table*}
    \centering
    \begin{tabular}{|m{2em} || m{16em}| m{16em}|}
        \toprule
        Class ($c$) & PGD-AT $\Lambda^{\mathit{adv\_diff\_org}}_{c}$ & STD-TR $\Lambda^{\mathit{adv\_diff\_org}}_{c}$ \\
        \midrule
        4 & \includegraphics[width=16em,keepaspectratio]{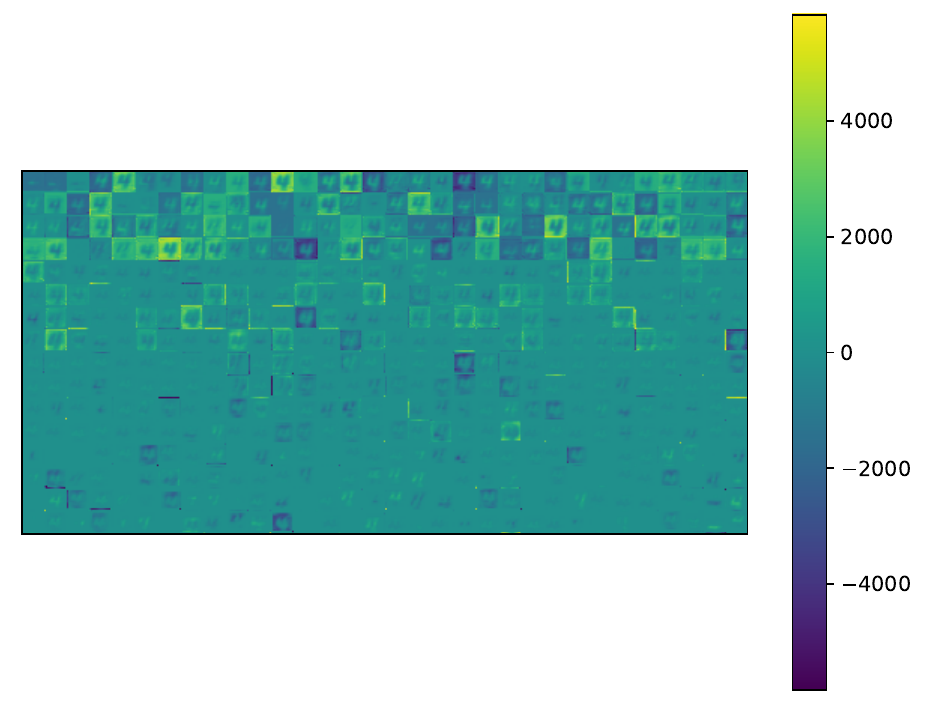} & \includegraphics[width=16em,keepaspectratio]{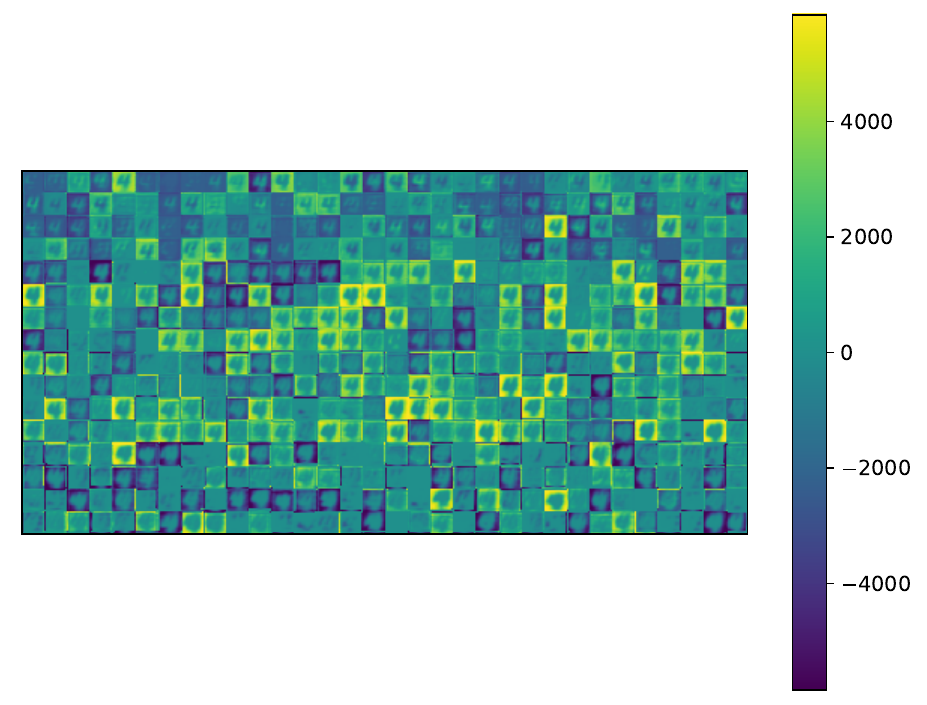}\\
        \hline
        5 & \includegraphics[width=16em,keepaspectratio]{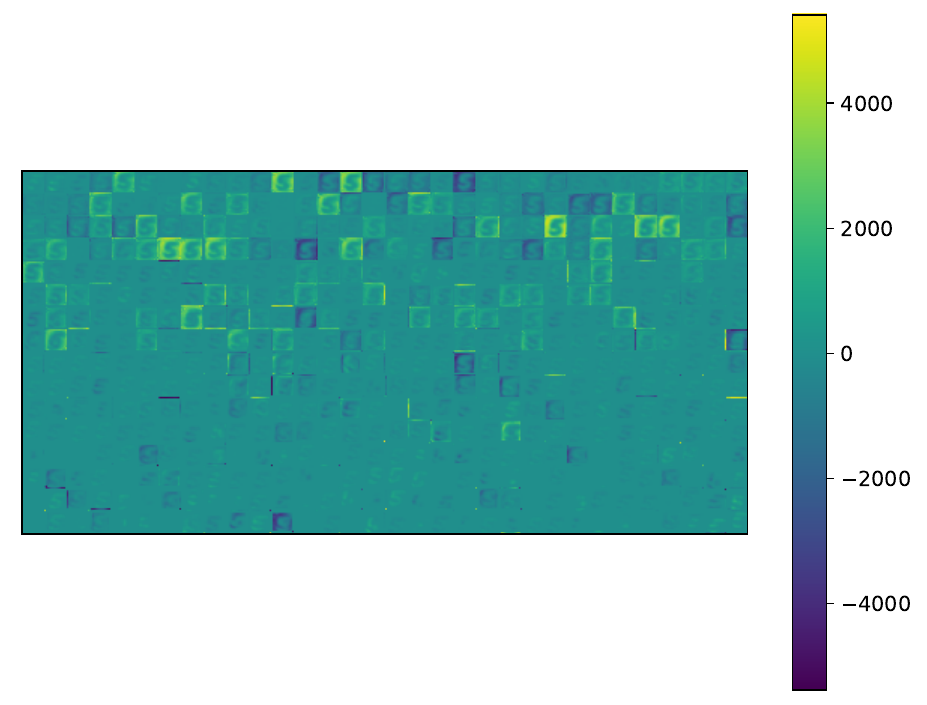} & \includegraphics[width=16em,keepaspectratio]{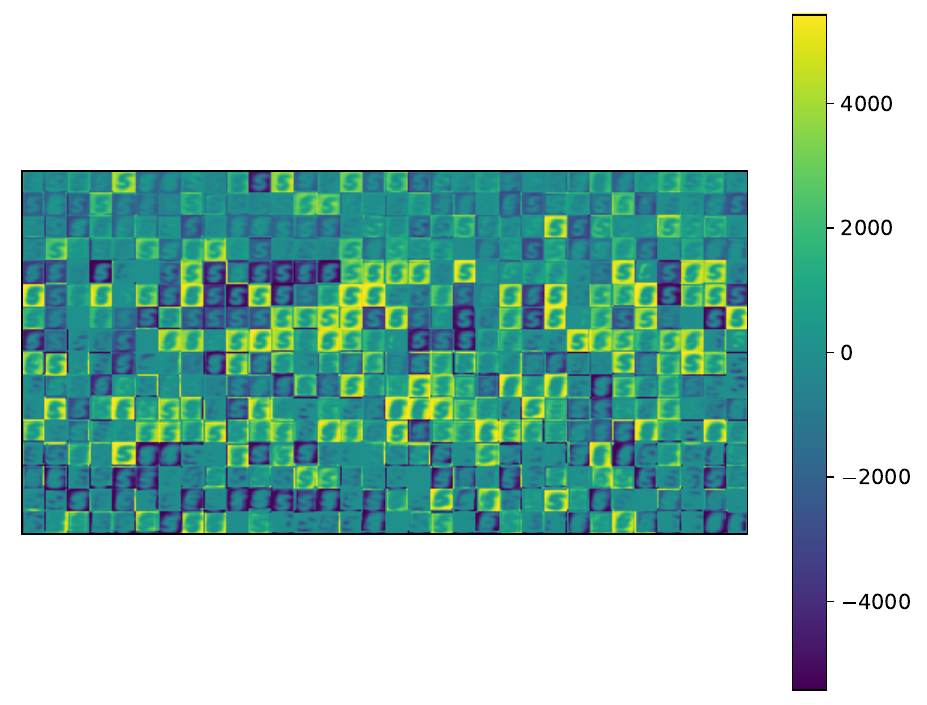}\\
        \hline
        6 & \includegraphics[width=16em,keepaspectratio]{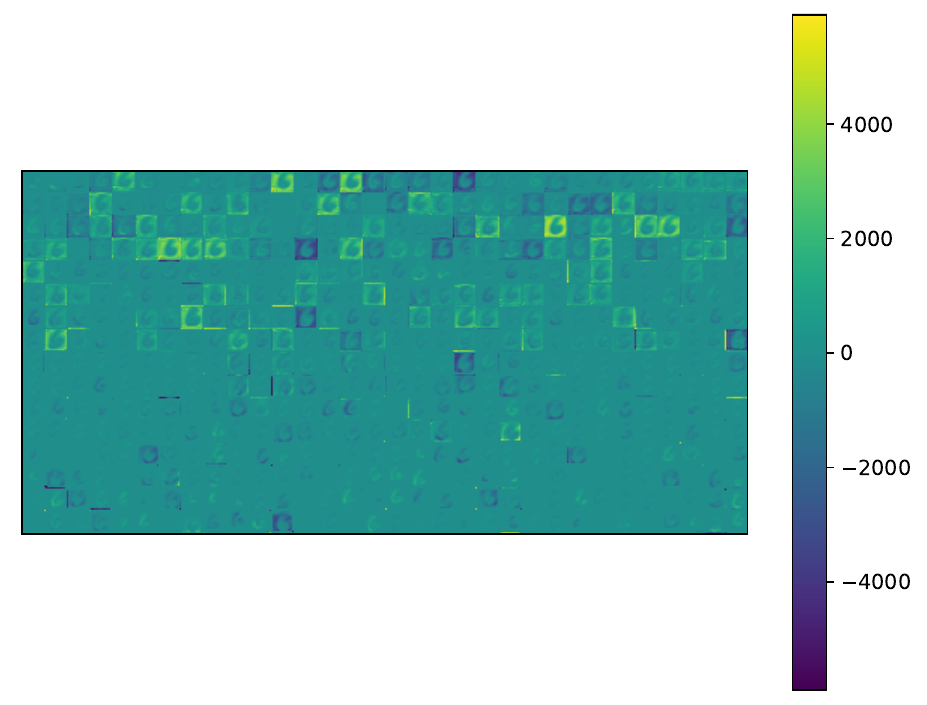} & \includegraphics[width=16em,keepaspectratio]{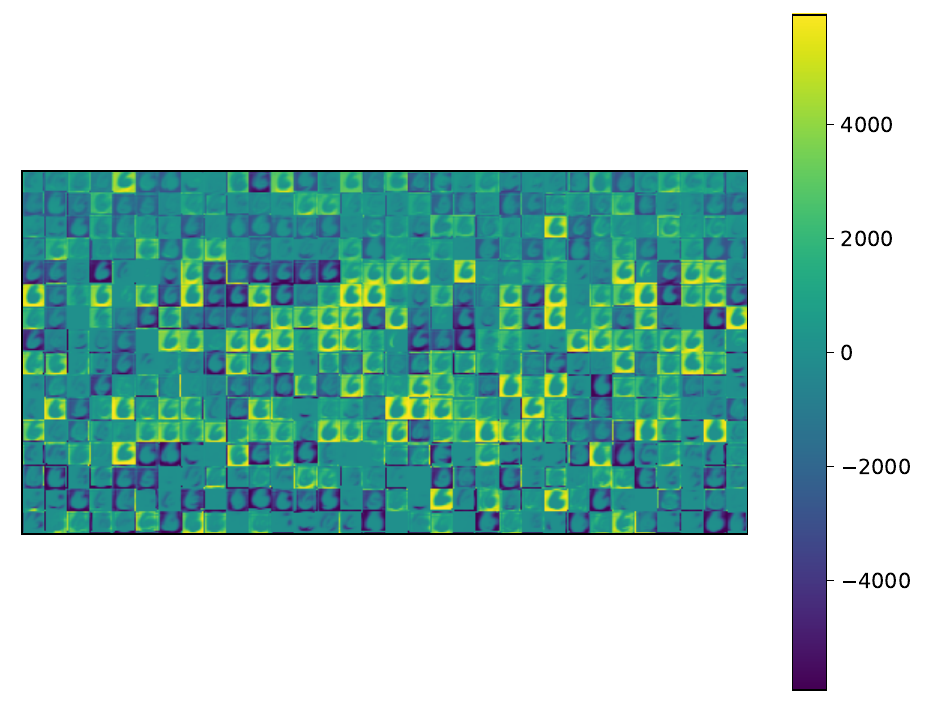}\\
        \hline
        7 & \includegraphics[width=16em,keepaspectratio]{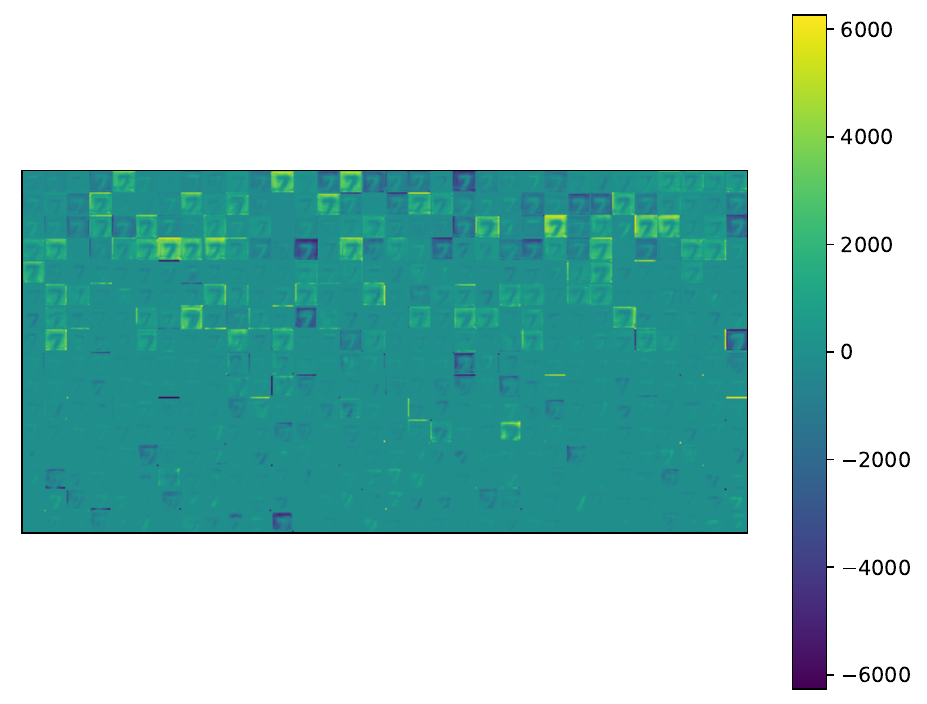} & \includegraphics[width=16em,keepaspectratio]{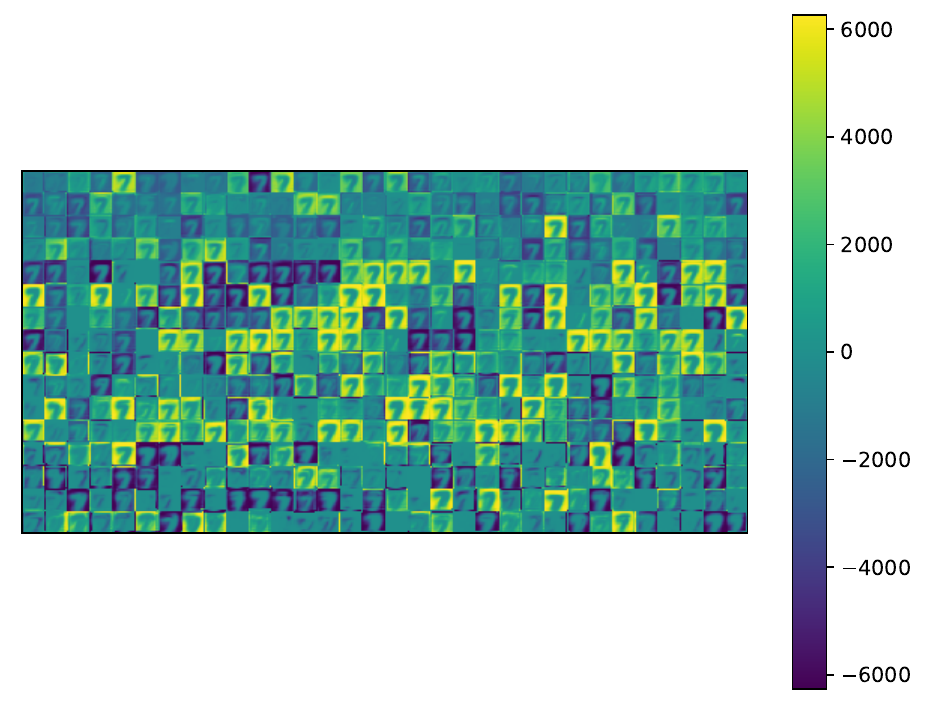}\\
        \bottomrule
    \end{tabular}
    \caption{$\Lambda^{\mathit{adv\_diff\_org}}_{c}$ for PGD-AT and STD-TR models with CONV\_N128\_D4 \dlgnmodel architecture on MNIST dataset. In each cell of the image, every four rows represent a layer's $\Lambda^{\mathit{adv\_diff\_org}}_{l,c}$}
    \label{tab:appnd_active_gate_count_adv_diff_org_part2}
\end{table*}

\subsection{Quantitative analysis of gating patterns in PGD-AT and STD-TR models} \label{subsec:appnd_gate_ovrlp}

The number of active gates per class at each pixel in \( F_{l} \) across all $L$ layers is given by \Cref{eq:active_gate_count_def}.
\begin{subequations}
    \begin{align}
        Gate(x) &=
    \begin{cases}
        1 , & \text{if  } x > 0 \\
        0 , & \text{otherwise}
    \end{cases}\\
    \Lambda^{mode}_{c} &= \sum_{i=1}^{N_{c}}Gate(F^{mode}(X_{c})) , &\in R^{L,C_{l},W,H} \label{eq:active_gate_count_def}    
    \end{align}
\end{subequations}
The following is the procedure to obtain IOU of active gate count of class $c_1$ and $c_2$ ($IOU_{agc}(c_1,c_2)$).

\begin{enumerate}
    \item Compute union of active gate counts at all pixels\\ 
    $A_{c_1,c_2}^{mode}(i)=\Lambda_{c_1}^{mode}(i)+\Lambda_{c_2}^{mode}(i) \; \forall i \in R^{L,C_l,W,H}$
    \item Compute intersection of active gate counts at all pixels\\
     $B_{c_1,c_2}^{mode}(i)=min(\Lambda_{c_1}^{mode}(i),\Lambda_{c_2}^{mode}(i)) \; \forall i \in R^{L,C_l,W,H}$
    \item Record the indices of $U_{c_1,c_2}^{mode}$ whose value is such that $A_{c_1,c_2}^{mode}(i) > 0.1*(|X_{c_1}|+|X_{c_2}|)$. Let such an index vector be $\iota_{c_1,c_2} \in R^{d}$. The intent of this stage is to remove outliers in the union of active gate counts of both classes.
    \item Obtain the final intersection as $I_{c_1,c_2}^{mode}=B_{c_1,c_2}[\iota_{c_1,c_2}^{mode}] \in R^{d}$. Obtain the final union region as $U_{c_1,c_2}^{mode}=A_{c_1,c_2}^{mode}[\iota_{c_1,c_2}^{mode}] \in R^{d}$
    \item Obtain overall average IOU between classes $c_1$,$c_2$ as\\ $IOU_{agc}^{mode}(c_1,c_2)=\frac{1}{d}\sum_{i=1}^{d}\frac{I_{c_1,c_2}^{mode}(i)}{U_{c_1,c_2}^{mode}(i)}$
\end{enumerate}

The $IOU_{agc}^{org}$, $IOU_{agc}^{adv}$ is measured for each pair of classes in \Cref{tab:appnd_iou_agc_class_fmnist} for Fashion MNIST dataset.

\begin{table*}
    \centering
    \begin{tabular}{|m{1.5em} | m{2em} | m{4em}|| m{2em} | m{2em}| m{2em} | m{2em}| m{2em} | m{2em}| m{2em} | m{2em}| m{2em} | m{2em}|}
        \toprule
        Src Class & Train Type & Quantity & Class 0 & Class 1 & Class 2 & Class 3 & Class 4 & Class 5 & Class 6 & Class 7 & Class 8 & Class 9 \\
        \midrule
        \multirow{4}{1.5em}{0} & \multirow{2}{2em}{PGD-AT} & $IOU^{adv}_{agc}$ & 100.0   & 81.0    & 86.9    & 86.4    & 86.0    & 75.0    & 89.6    & 73.5    & 80.9    & 75.6    \\
        & & $IOU^{org}_{agc}$ & 100.0   & 75.6    & 84.3    & 83.0    & 80.9    & 68.1    & 86.5    & 67.4    & 74.9    & 69.4    \\ \cline{2-13}
        & \multirow{2}{2em}{STD-TR} & $IOU^{adv}_{agc}$ & 100.0   & 72.1    & 78.6    & 79.4    & 76.9    & 67.9    & 83.2    & 65.3    & 73.9    & 67.7    \\
        & & $IOU^{org}_{agc}$ & 100.0   & 57.1    & 69.2    & 69.9    & 65.4    & 47.3    & 76.6    & 44.4    & 59.6    & 49.4    \\ \hline
        \multirow{4}{1.5em}{1} & \multirow{2}{2em}{PGD-AT} & $IOU^{adv}_{agc}$ & 81.0    & 100.0   & 77.8    & 88.2    & 80.2    & 74.2    & 77.8    & 74.4    & 74.0    & 73.6    \\
        & & $IOU^{org}_{agc}$ & 75.6    & 100.0   & 72.1    & 85.2    & 74.5    & 66.2    & 71.6    & 67.5    & 65.9    & 66.4    \\ \cline{2-13}
        & \multirow{2}{2em}{STD-TR} & $IOU^{adv}_{agc}$ & 72.1    & 100.0   & 68.5    & 82.4    & 71.1    & 64.9    & 70.9    & 65.5    & 66.7    & 64.2    \\
        & & $IOU^{org}_{agc}$ & 57.1    & 100.0   & 51.0    & 72.8    & 55.1    & 45.9    & 52.1    & 44.9    & 46.8    & 45.4    \\ \hline
        \multirow{4}{1.5em}{2} & \multirow{2}{2em}{PGD-AT} & $IOU^{adv}_{agc}$ & 86.9    & 77.8    & 100.0   & 82.3    & 91.3    & 76.0    & 93.1    & 73.5    & 84.8    & 77.5    \\
        & & $IOU^{org}_{agc}$ & 84.3    & 72.1    & 100.0   & 77.4    & 89.8    & 69.2    & 91.4    & 67.2    & 78.8    & 72.6    \\ \cline{2-13}
        & \multirow{2}{2em}{STD-TR} & $IOU^{adv}_{agc}$ & 78.6    & 68.5    & 100.0   & 73.1    & 86.3    & 67.0    & 83.7    & 65.4    & 76.9    & 68.8    \\
        & & $IOU^{org}_{agc}$ & 69.2    & 51.0    & 100.0   & 58.4    & 81.0    & 48.5    & 81.9    & 45.3    & 65.2    & 54.4    \\ \hline
        \multirow{4}{1.5em}{3} & \multirow{2}{2em}{PGD-AT} & $IOU^{adv}_{agc}$ & 86.4    & 88.2    & 82.3    & 100.0   & 84.7    & 75.0    & 83.0    & 74.7    & 77.7    & 75.2    \\
        & & $IOU^{org}_{agc}$ & 83.0    & 85.2    & 77.4    & 100.0   & 79.7    & 67.7    & 78.0    & 68.3    & 70.8    & 68.5    \\ \cline{2-13}
        & \multirow{2}{2em}{STD-TR} & $IOU^{adv}_{agc}$ & 79.4    & 82.4    & 73.1    & 100.0   & 75.4    & 68.4    & 77.5    & 68.2    & 71.6    & 68.3    \\
        & & $IOU^{org}_{agc}$ & 69.9    & 72.8    & 58.4    & 100.0   & 62.2    & 49.4    & 62.8    & 47.2    & 54.7    & 49.4    \\ \hline
        \multirow{4}{1.5em}{4} & \multirow{2}{2em}{PGD-AT} & $IOU^{adv}_{agc}$ & 86.0    & 80.2    & 91.3    & 84.7    & 100.0   & 76.1    & 91.3    & 74.4    & 84.1    & 78.2    \\
        & & $IOU^{org}_{agc}$ & 80.9    & 74.5    & 89.8    & 79.7    & 100.0   & 68.0    & 87.9    & 67.1    & 78.0    & 72.5    \\ \cline{2-13}
        & \multirow{2}{2em}{STD-TR} & $IOU^{adv}_{agc}$ & 76.9    & 71.1    & 86.3    & 75.4    & 100.0   & 66.4    & 83.5    & 65.4    & 76.1    & 69.0    \\
        & & $IOU^{org}_{agc}$ & 65.4    & 55.1    & 81.0    & 62.2    & 100.0   & 48.9    & 79.4    & 46.0    & 64.9    & 54.8    \\
        \bottomrule
    \end{tabular}
    \caption{CONV \dlgnmodel-N128-D4 PGD-AT vs STD-TR model IOU of active gate count between class-pairs over adversarial and original examples for Fashion MNIST dataset.}
    \label{tab:appnd_iou_agc_class_fmnist}
\end{table*}

\begin{table*}
    \centering
    \begin{tabular}{|m{1.5em} | m{2em} | m{4em}|| m{2em} | m{2em}| m{2em} | m{2em}| m{2em} | m{2em}| m{2em} | m{2em}| m{2em} | m{2em}|}
        \toprule
        Src Class & Train Type & Quantity & Class 0 & Class 1 & Class 2 & Class 3 & Class 4 & Class 5 & Class 6 & Class 7 & Class 8 & Class 9 \\
        \midrule
        \multirow{4}{1.5em}{0} & \multirow{2}{2em}{PGD-AT} & $IOU^{adv}_{agc}$ & 100 & 70.2 & 83 & 82.7 & 81.8 & 83.9 & 82.7 & 77.4 & 84 & 80.6 \\
        & & $IOU^{org}_{agc}$ & 100 & 66.2 & 79.3 & 79.4 & 77.9 & 81.3 & 78.5 & 72.7 & 81.2 & 76 \\ \cline{2-13}
        & \multirow{2}{2em}{STD-TR} & $IOU^{adv}_{agc}$ & 100 & 78.1 & 84.7 & 82 & 81 & 84.5 & 84 & 80 & 78.2 & 82.6 \\
        & & $IOU^{org}_{agc}$ &  100 & 59.7 & 74.7 & 75 & 73.1 & 77.6 & 73.2 & 66.2 & 77.5 & 69.7 \\
        \hline
        \multirow{4}{1.5em}{1} & \multirow{2}{2em}{PGD-AT} & $IOU^{adv}_{agc}$ & 70.2 & 100 & 74.9 & 75.8 & 74.7 & 74.2 & 75 & 77.2 & 76.3 & 75.6 \\
        & & $IOU^{org}_{agc}$ & 66.2 & 100 & 71.9 & 74.3 & 71.9 & 71 & 71.6 & 75 & 74.2 & 73.6 \\ \cline{2-13}
        & \multirow{2}{2em}{STD-TR} & $IOU^{adv}_{agc}$ & 78.1 & 100 & 82.7 & 79.5 & 80 & 80 & 79.3 & 83.3 & 74.9 & 80.4 \\
        & & $IOU^{org}_{agc}$ & 59.7 & 100 & 63.7 & 66.5 & 65.6 & 64.4 & 64.7 & 67.4 & 68.4 & 67.2 \\
        \hline
        \multirow{4}{1.5em}{2} & \multirow{2}{2em}{PGD-AT} & $IOU^{adv}_{agc}$ & 83 & 74.9 & 100 & 86.9 & 84.3 & 83.5 & 85.8 & 79.7 & 86.1 & 82.4 \\
        & & $IOU^{org}_{agc}$ & 79.3 & 71.9 & 100 & 84.7 & 80.5 & 80.1 & 82.7 & 74.9 & 83.4 & 78.3 \\ \cline{2-13}
        & \multirow{2}{2em}{STD-TR} & $IOU^{adv}_{agc}$ & 84.7 & 82.7 & 100 & 82.4 & 83.9 & 83.3 & 85.7 & 82 & 79.2 & 83 \\
        & & $IOU^{org}_{agc}$ & 74.7 & 63.7 & 100 & 80.7 & 74.2 & 74.9 & 77 & 68.2 & 77.5 & 70.5 \\
        \hline
        \multirow{4}{1.5em}{3} & \multirow{2}{2em}{PGD-AT} & $IOU^{adv}_{agc}$ & 82.8 & 75.8 & 86.9 & 100 & 82.7 & 86.3 & 82.1 & 81.7 & 87.4 & 83 \\
        & & $IOU^{org}_{agc}$ & 79.4 & 74.3 & 84.7 & 100 & 78.6 & 84.7 & 78 & 77.8 & 85.4 & 79 \\ \cline{2-13}
        & \multirow{2}{2em}{STD-TR} & $IOU^{adv}_{agc}$ & 82 & 79.5 & 82.4 & 100 & 77.5 & 85.4 & 77.6 & 83.6 & 75.7 & 81.8 \\
        & & $IOU^{org}_{agc}$ & 75 & 66.5 & 80.7 & 100 & 73.4 & 80.8 & 72.3 & 71.9 & 81.4 & 73 \\
        \hline
        \multirow{4}{1.5em}{4} & \multirow{2}{2em}{PGD-AT} & $IOU^{adv}_{agc}$ & 81.8 & 74.7 & 84.3 & 82.7 & 100 & 85.8 & 84.7 & 85.4 & 86.9 & 91.1 \\
        & & $IOU^{org}_{agc}$ & 77.9 & 71.9 & 80.5 & 78.6 & 100 & 82.3 & 81.3 & 82.8 & 84.5 & 90.6 \\ \cline{2-13}
        & \multirow{2}{2em}{STD-TR} & $IOU^{adv}_{agc}$ & 81 & 80.2 & 83.9 & 77.8 & 100 & 81.2 & 82.5 & 80.7 & 80.8 & 87.2 \\
        & & $IOU^{org}_{agc}$ & 73.1 & 65.6 & 74.2 & 73.4 & 100 & 78.5 & 75 & 77.7 & 80 & 85.9 \\
        \bottomrule
    \end{tabular}
    \caption{CONV \dlgnmodel-N128-D4 PGD-AT vs STD-TR model IOU of active gate count between class-pairs over adversarial and original examples in MNIST dataset.}
    \label{tab:appnd_iou_agc_class_mnist}
\end{table*}

\subsection{Interpretation of gating patterns in PGD-AT vs STD-TR models}\label{subsec:appnd_visualization_pgdat_stdtr}

The visualizations (\(I^{org}\),\(I^{adv}\))  for CONV-\dlgnmodel\_N128\_D4 trained on the MNIST, Fashion MNIST dataset are presented in \Cref{tab:vis_adv_org_pgdat_stdtr}. We report visualized images $I^{ado}_{c},I^{amo}_{c}$ for both CONV-\dlgnmodel\_N128\_D4 PGD-AT and STD-TR models as before trained on MNIST, Fashion MNIST dataset in \Cref{tab:vis_adv_and_org_pgdat_stdtr}.

\begin{table*}
    \centering
    \begin{tabular}{|m{2em} | m{3em}| m{3em} || m{3em} | m{3em}||m{2.6em} | m{3em}| m{3em} || m{3em} | m{3em}|}
        \toprule
        MN Class ($c$) & PGD-AT $I^{org}$ & PGD-AT $I^{adv}$ & STD-TR $I^{org}$ & STD-TR $I^{adv}$ & Fashion MN ($c$) & PGD-AT $I^{org}$ & PGD-AT $I^{adv}$ & STD-TR $I^{org}$ & STD-TR $I^{adv}$ \\
        \midrule
        0 & \includegraphics[width=3em,keepaspectratio]{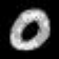} & \includegraphics[width=3em,keepaspectratio]{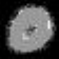} & \includegraphics[width=3em,keepaspectratio]{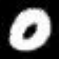} & \includegraphics[width=3em,keepaspectratio]{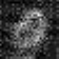} &
        Ankle- boot & \includegraphics[width=3em,keepaspectratio]{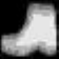} & \includegraphics[width=3em,keepaspectratio]{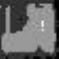} & \includegraphics[width=3em,keepaspectratio]{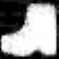} & \includegraphics[width=3em,keepaspectratio]{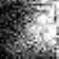}\\
        \hline
        1 & \includegraphics[width=3em,keepaspectratio]{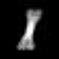} & \includegraphics[width=3em,keepaspectratio]{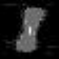} & \includegraphics[width=3em,keepaspectratio]{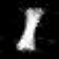} & \includegraphics[width=3em,keepaspectratio]{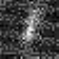} &
        Bag & \includegraphics[width=3em,keepaspectratio]{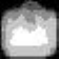} & \includegraphics[width=3em,keepaspectratio]{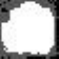} & \includegraphics[width=3em,keepaspectratio]{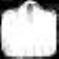} & \includegraphics[width=3em,keepaspectratio]{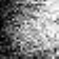}\\
        \hline
        2 & \includegraphics[width=3em,keepaspectratio]{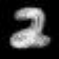} & \includegraphics[width=3em,keepaspectratio]{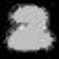} & \includegraphics[width=3em,keepaspectratio]{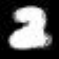} & \includegraphics[width=3em,keepaspectratio]{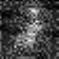} &
        Coat & \includegraphics[width=3em,keepaspectratio]{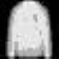} & \includegraphics[width=3em,keepaspectratio]{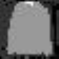} & \includegraphics[width=3em,keepaspectratio]{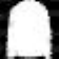} & \includegraphics[width=3em,keepaspectratio]{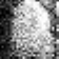}\\
        \hline
        3 & \includegraphics[width=3em,keepaspectratio]{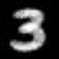} & \includegraphics[width=3em,keepaspectratio]{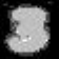} & \includegraphics[width=3em,keepaspectratio]{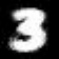} & \includegraphics[width=3em,keepaspectratio]{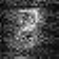} &
        Dress & \includegraphics[width=3em,keepaspectratio]{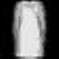} & \includegraphics[width=3em,keepaspectratio]{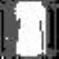} & \includegraphics[width=3em,keepaspectratio]{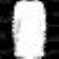} & \includegraphics[width=3em,keepaspectratio]{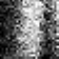}\\
        \hline
        4 & \includegraphics[width=3em,keepaspectratio]{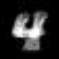} & \includegraphics[width=3em,keepaspectratio]{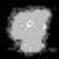} & \includegraphics[width=3em,keepaspectratio]{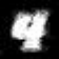} & \includegraphics[width=3em,keepaspectratio]{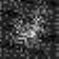} &
        Pullover & \includegraphics[width=3em,keepaspectratio]{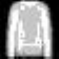} & \includegraphics[width=3em,keepaspectratio]{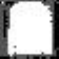} & \includegraphics[width=3em,keepaspectratio]{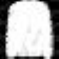} & \includegraphics[width=3em,keepaspectratio]{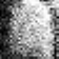}\\
        \hline
        5 & \includegraphics[width=3em,keepaspectratio]{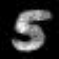} & \includegraphics[width=3em,keepaspectratio]{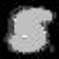} & \includegraphics[width=3em,keepaspectratio]{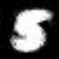} & \includegraphics[width=3em,keepaspectratio]{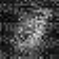} &
        Sandal & \includegraphics[width=3em,keepaspectratio]{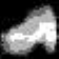} & \includegraphics[width=3em,keepaspectratio]{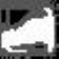} & \includegraphics[width=3em,keepaspectratio]{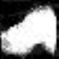} & \includegraphics[width=3em,keepaspectratio]{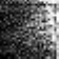}\\
        \hline
        6 & \includegraphics[width=3em,keepaspectratio]{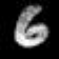} & \includegraphics[width=3em,keepaspectratio]{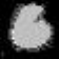} & \includegraphics[width=3em,keepaspectratio]{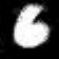} & \includegraphics[width=3em,keepaspectratio]{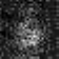} & 
        Shirt & \includegraphics[width=3em,keepaspectratio]{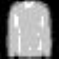} & \includegraphics[width=3em,keepaspectratio]{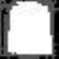} & \includegraphics[width=3em,keepaspectratio]{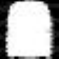} & \includegraphics[width=3em,keepaspectratio]{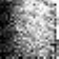}\\
        \hline
        7 & \includegraphics[width=3em,keepaspectratio]{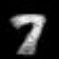} & \includegraphics[width=3em,keepaspectratio]{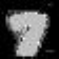} & \includegraphics[width=3em,keepaspectratio]{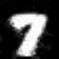} & \includegraphics[width=3em,keepaspectratio]{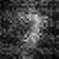} &
        Sneaker & \includegraphics[width=3em,keepaspectratio]{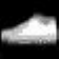} & \includegraphics[width=3em,keepaspectratio]{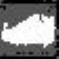} & \includegraphics[width=3em,keepaspectratio]{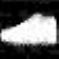} & \includegraphics[width=3em,keepaspectratio]{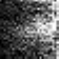}\\
        \hline
        8 & \includegraphics[width=3em,keepaspectratio]{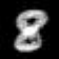} & \includegraphics[width=3em,keepaspectratio]{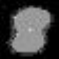} & \includegraphics[width=3em,keepaspectratio]{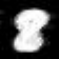} & \includegraphics[width=3em,keepaspectratio]{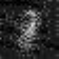} &
        T-shirt & \includegraphics[width=3em,keepaspectratio]{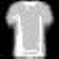} & \includegraphics[width=3em,keepaspectratio]{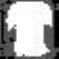} & \includegraphics[width=3em,keepaspectratio]{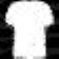} & \includegraphics[width=3em,keepaspectratio]{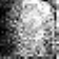}\\
        \hline
        9 & \includegraphics[width=3em,keepaspectratio]{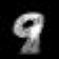} & \includegraphics[width=3em,keepaspectratio]{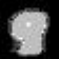} & \includegraphics[width=3em,keepaspectratio]{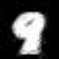} & \includegraphics[width=3em,keepaspectratio]{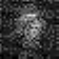} &
        Trouser & \includegraphics[width=3em,keepaspectratio]{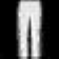} & \includegraphics[width=3em,keepaspectratio]{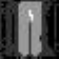} & \includegraphics[width=3em,keepaspectratio]{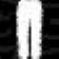} & \includegraphics[width=3em,keepaspectratio]{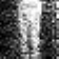}\\
        \bottomrule
    \end{tabular}
    \caption{Image $I$ which triggers dominating gating pattern per class obtained on adversarial examples (column 3,5,8,10) and original examples (column 2,4,7,9). Columns 2,3,7,8 are on the PGD-AT model, and columns 4,5,9,10 are on the STD-TR model. Loss function is as per \Cref{eq:vis_tanh_loss_def},$\lambda=0.9,\alpha=0.1$,optimization is as per \Cref{eq:vis_signed_grad}}
    \label{tab:vis_adv_org_pgdat_stdtr}
\end{table*}

\begin{table*}
    \centering
    \begin{tabular}{|m{2em} | m{3em}| m{3em} || m{3em} | m{3em}||m{2.6em} | m{3em}| m{3em} || m{3em} | m{3em}|}
        \toprule
        MN Class ($c$) & PGD-AT $I^{ado}$ & PGD-AT $I^{amo}$ & STD-TR $I^{ado}$ & STD-TR $I^{amo}$ & Fashion MN ($c$) & PGD-AT $I^{ado}$ & PGD-AT $I^{amo}$ & STD-TR $I^{ado}$ & STD-TR $I^{amo}$ \\
        \midrule
        0 & \includegraphics[width=3em,keepaspectratio]{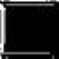} & \includegraphics[width=3em,keepaspectratio]{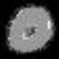} & \includegraphics[width=3em,keepaspectratio]{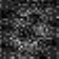} & \includegraphics[width=3em,keepaspectratio]{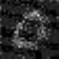} &
        Ankle-boot & \includegraphics[width=3em,keepaspectratio]{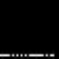} & \includegraphics[width=3em,keepaspectratio]{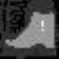} & \includegraphics[width=3em,keepaspectratio]{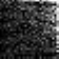} & \includegraphics[width=3em,keepaspectratio]{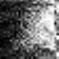}\\
        \hline
        1 & \includegraphics[width=3em,keepaspectratio]{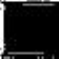} & \includegraphics[width=3em,keepaspectratio]{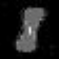} & \includegraphics[width=3em,keepaspectratio]{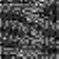} & \includegraphics[width=3em,keepaspectratio]{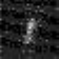} &
        Bag & \includegraphics[width=3em,keepaspectratio]{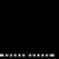} & \includegraphics[width=3em,keepaspectratio]{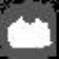} & \includegraphics[width=3em,keepaspectratio]{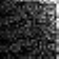} & \includegraphics[width=3em,keepaspectratio]{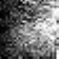}\\
        \hline
        2 & \includegraphics[width=3em,keepaspectratio]{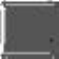} & \includegraphics[width=3em,keepaspectratio]{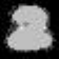} & \includegraphics[width=3em,keepaspectratio]{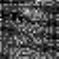} & \includegraphics[width=3em,keepaspectratio]{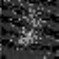} &
        Coat & \includegraphics[width=3em,keepaspectratio]{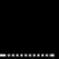} & \includegraphics[width=3em,keepaspectratio]{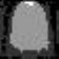} & \includegraphics[width=3em,keepaspectratio]{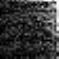} & \includegraphics[width=3em,keepaspectratio]{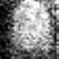}\\
        \hline
        3 & \includegraphics[width=3em,keepaspectratio]{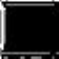} & \includegraphics[width=3em,keepaspectratio]{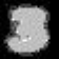} & \includegraphics[width=3em,keepaspectratio]{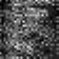} & \includegraphics[width=3em,keepaspectratio]{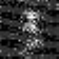} & 
        Dress & \includegraphics[width=3em,keepaspectratio]{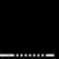} & \includegraphics[width=3em,keepaspectratio]{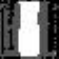} & \includegraphics[width=3em,keepaspectratio]{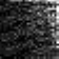} & \includegraphics[width=3em,keepaspectratio]{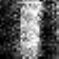}\\
        \hline
        4 & \includegraphics[width=3em,keepaspectratio]{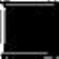} & \includegraphics[width=3em,keepaspectratio]{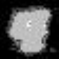} & \includegraphics[width=3em,keepaspectratio]{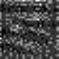} & \includegraphics[width=3em,keepaspectratio]{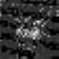} &
        Pullover & \includegraphics[width=3em,keepaspectratio]{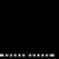} & \includegraphics[width=3em,keepaspectratio]{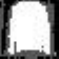} & \includegraphics[width=3em,keepaspectratio]{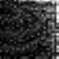} & \includegraphics[width=3em,keepaspectratio]{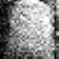}\\
        \hline
        5 & \includegraphics[width=3em,keepaspectratio]{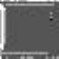} & \includegraphics[width=3em,keepaspectratio]{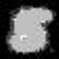} & \includegraphics[width=3em,keepaspectratio]{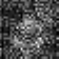} & \includegraphics[width=3em,keepaspectratio]{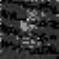} &
        Sandal & \includegraphics[width=3em,keepaspectratio]{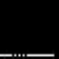} & \includegraphics[width=3em,keepaspectratio]{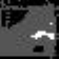} & \includegraphics[width=3em,keepaspectratio]{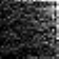} & \includegraphics[width=3em,keepaspectratio]{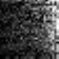}\\
        \hline
        6 & \includegraphics[width=3em,keepaspectratio]{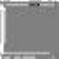} & \includegraphics[width=3em,keepaspectratio]{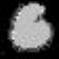} & \includegraphics[width=3em,keepaspectratio]{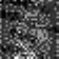} & \includegraphics[width=3em,keepaspectratio]{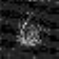} &
        Shirt & \includegraphics[width=3em,keepaspectratio]{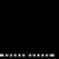} & \includegraphics[width=3em,keepaspectratio]{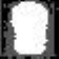} & \includegraphics[width=3em,keepaspectratio]{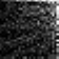} & \includegraphics[width=3em,keepaspectratio]{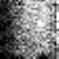}\\
        \hline
        7 & \includegraphics[width=3em,keepaspectratio]{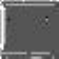} & \includegraphics[width=3em,keepaspectratio]{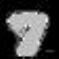} & \includegraphics[width=3em,keepaspectratio]{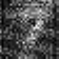} & \includegraphics[width=3em,keepaspectratio]{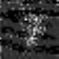} &
        Sneaker & \includegraphics[width=3em,keepaspectratio]{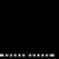} & \includegraphics[width=3em,keepaspectratio]{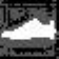} & \includegraphics[width=3em,keepaspectratio]{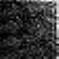} & \includegraphics[width=3em,keepaspectratio]{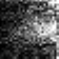}\\
        \hline
        8 & \includegraphics[width=3em,keepaspectratio]{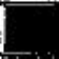} & \includegraphics[width=3em,keepaspectratio]{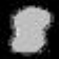} & \includegraphics[width=3em,keepaspectratio]{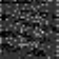} & \includegraphics[width=3em,keepaspectratio]{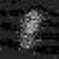} &
        T-shirt & \includegraphics[width=3em,keepaspectratio]{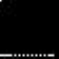} & \includegraphics[width=3em,keepaspectratio]{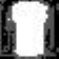} & \includegraphics[width=3em,keepaspectratio]{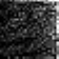} & \includegraphics[width=3em,keepaspectratio]{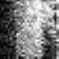}\\
        \hline
        9 & \includegraphics[width=3em,keepaspectratio]{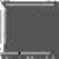} & \includegraphics[width=3em,keepaspectratio]{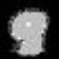} & \includegraphics[width=3em,keepaspectratio]{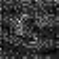} & \includegraphics[width=3em,keepaspectratio]{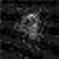} &
        Trouser & \includegraphics[width=3em,keepaspectratio]{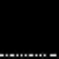} & \includegraphics[width=3em,keepaspectratio]{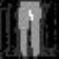} & \includegraphics[width=3em,keepaspectratio]{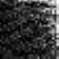} & \includegraphics[width=3em,keepaspectratio]{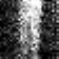}\\
        \bottomrule
    \end{tabular}
    \caption{Image $I$ which triggers dominating active gating pattern per class obtained on adversarial examples alone but not on original examples (columns 2,4,7,9) and obtained both on original examples and adversarial examples(columns 3,5,8,10). Columns 2,3,7,8 are on the PGD-AT model, and columns 4,5,9,10 are on the STD-TR models. Loss function is as per \Cref{eq:vis_tanh_loss_def},$\lambda=0.9,\alpha=0.1$,optimization is as per \Cref{eq:vis_signed_grad}}
    \label{tab:vis_adv_and_org_pgdat_stdtr}
\end{table*}


\end{document}